\documentclass[lettersize,journal]{IEEEtran}

\usepackage{graphicx}
\makeatletter
\def\input@path{{content/}{meta/}{./}}
\graphicspath{{figures/}{figures/architecture}{figures/appendix}{./}{figures/ECG5000/}{figures/preterm/}{figures/wrap/}{authors/}} 
\makeatother

\newcommand{\system}{SigTime}
\title{\system: Learning and Visually Explaining Time Series Signatures}
\author{
Yu-Chia Huang, 
Juntong Chen, 
Dongyu Liu, 
and Kwan-Liu Ma, \textit{Fellow, IEEE}
\thanks{Yu-Chia Huang, Juntong Chen, Dongyu Liu and Kwan-Liu Ma are with the Department of Computer Science, University of California, Davis, One Shields Ave., Davis, California 95616. E-mail: ycihuang, jtochen, dyuliu, klma@ucdavis.edu}
}

\usepackage{amsmath,amsfonts}
\usepackage{algorithmic}
\usepackage{algorithm}
\usepackage{array}
\usepackage[caption=false,font=normalsize,labelfont=sf,textfont=sf]{subfig}
\usepackage{textcomp}
\usepackage{stfloats}
\usepackage{url}
\usepackage{verbatim}
\usepackage{graphicx}
\usepackage{cite}
\usepackage{enumitem}
\usepackage{amssymb}
\usepackage{enumitem}
\usepackage{graphicx}
\usepackage{wrapfig}
\usepackage{tabularx}
\usepackage[dvipsnames]{xcolor}
\usepackage{tcolorbox}
\usepackage{marginnote}
\usepackage{adjustbox}
\usepackage{hyperref}
\usepackage{pgfplots}
\usepackage{multirow}

\definecolor{dypink}{HTML}{ec008c}

\usepackage[color=purple!15]{todonotes}

\definecolor{lightblue}{RGB}{199, 233, 255}

\newcommand{\req}[1]{\tcbox[on line, size=tight, boxrule=0pt, colback=lightblue!100,left=3pt,right=3pt,top=2pt,bottom=2pt, arc=5pt, frame empty ]{\textcolor{black}{\textbf{#1}}}}

\global\marginparsep=3pt

\newcommand{\sidecomment}[1]{%
  \ifdefined\revise
  \marginnote{%
    \textcolor{Magenta}{%
      \adjustbox{minipage=0.65\marginparwidth,fbox}{%
          \scriptsize#1%
      }
    }
  }
  \fi
}

\newcommand{\revision}[1]{%
  \ifdefined\revise
  {\color{Magenta}#1}%
  \else
  #1%
  \fi
}

\begin{document}

\maketitle
\begin{abstract}
    
    Understanding and distinguishing temporal patterns in time series data is essential for scientific discovery and decision-making. 
    For example, in biomedical research, uncovering meaningful patterns in physiological signals can improve diagnosis, risk assessment, and patient outcomes.
    However, existing methods for time series pattern discovery face major challenges, including high computational complexity, limited interpretability, and difficulty in capturing meaningful temporal structures.
    To address these gaps, we introduce a novel learning framework that jointly trains two Transformer models using complementary time series representations: shapelet-based representations to capture localized temporal structures and traditional feature engineering to encode statistical properties. 
    The learned shapelets serve as interpretable signatures that differentiate time series across classification labels.
    Additionally, we develop a visual analytics system---\system---with coordinated views to facilitate exploration of time series signatures from multiple perspectives, aiding in useful insights generation. 
    We quantitatively evaluate our learning framework on eight publicly available datasets and one proprietary clinical dataset. Additionally, we demonstrate the effectiveness of our system through two usage scenarios along with the domain experts: one involving public ECG data and the other focused on preterm labor analysis.
\end{abstract}


\section{Introduction}
Identifying temporal characteristics that differentiate time series across distinct classes is a central problem in time series analysis~\cite{mitsa2010temporal,maharaj2019time,gupta2020approaches}. These temporal patterns are often the primary basis for classification in domains such as healthcare, human behavior, and transportation. Accurately identifying them not only improves classification performance but also supports insight discovery and assists in decision-making.
For example, morphological (shape-based or structural) differences in Electrocardiogram (ECG) signals support the detection of cardiac anomalies~\cite{olszewski2001generalized,maharaj2014discriminant};
motion and rhythm patterns enable the recognition of human activities~\cite{seto2015multivariate} and speech signals ~\cite{gee2019explaining}.

Despite its importance, time series classification remains challenging due to high dimensionality, complex temporal dependencies, and the presence of intra- and inter-class variability and noise~\cite{maharaj2019time}. 
These factors make it difficult to extract temporal structures that are useful for distinguishing between classes~\cite{schafer2015boss}.
In addition, temporal patterns often evolve over time, requiring models that can adapt to non-stationary behavior. 
Traditional methods such as motif discovery~\cite{torkamani2017survey}, discriminative subsequence mining~\cite{ye2009time}, and symbolic representations~\cite{baydogan2015learning} have been used to extract informative, local structural patterns. 
While effective in some settings, these methods often require domain-specific parameter tuning and may not scale well computationally. 
More recent deep learning approaches~\cite{gee2019explaining, theissler2022explainable} automatically learn complex features and scale efficiently, but their limited interpretability restricts their use in domains where explanations are critical.

To address these challenges, we introduce a new learning framework that balances accuracy, interpretability, and computational scalability. Our approach jointly trains two Transformer models~\cite{vaswani2017attention} on complementary time series representations: (1) shapelet-based features that capture localized structural patterns, and (2) engineered features that summarize statistical characteristics. The shapelets~\cite{ye2009time}, which in this work we refer to as \textbf{Time Series Signatures}, are discriminative subsequences that highlight class-specific behaviors and provide interpretable signals tied to classification outcomes. 
These signatures focus on short, informative regions rather than the entire time series, offering robustness to noise and temporal misalignment. In domains such as cardiology or maternal health, where transparency supports expert decision-making, this interpretability is critical.


To support interactive analysis and further enhance the interpretability, we also develop an application-agnostic visual analytics system, \textbf{\system}, for exploring learned time series signatures. The design of the system was informed by a review of prior work, our own research in time series analysis and visualization, and ongoing collaboration with medical experts over two years.
The visual analysis tasks supported by the system reflect common analytical needs identified in both literature and practice, and were refined through regular feedback from collaborators.
Our system includes observations ranging from a broad overview and global feature exploration to a detailed view of individual elements. This enables the user to not only discover differences between various classes and clusters but also access detailed information about each.


In summary, our contributions are as follows:
\begin{itemize}
    \item We introduce a joint learning framework that combines shapelet-based and statistical representations of time series. The framework balances classification accuracy, interpretability, and computational scalability by learning Time Series Signatures through dual Transformer models.
    
    \item 
    We develop \system, a visual analytics system that supports interactive exploration of learned time series signatures across different analytical perspectives, from global feature patterns to fine-grained subsequence inspection. 
    Our implementation is available as open source on GitHub.\footnote{\url{https://github.com/via-cs/sigtime}}.

    \item  We conduct a comprehensive evaluation of our method on eight public time series datasets, demonstrating competitive classification performance. Additionally, we validate \system~through two usage scenarios focused on ECG data and preterm labor analysis.

\end{itemize}


\section{Related Works}

\subsection{Time Series Pattern Recognition}

Understanding behaviors and patterns in time series has been widely studied, with three main methodological approaches emerging~\cite{bagnall2017great}. 
The most straightforward approach is to \textbf{extract statistical summaries}---such as mean, variance, and extrema---over fixed intervals, representing each time series as a feature vector~\cite{christ2018time}. These representations are easy to compute and interpret but often cannot define the fine-grained temporal structure necessary to distinguish subtle or localized patterns, which can lead to sub-optimal accuracy.

The second category includes \textbf{structural pattern mining} techniques, such as motif discovery~\cite{torkamani2017survey}, discriminative subsequence mining (i.e., shapelets)~\cite{ye2009time}, and symbolic representations~\cite{baydogan2015learning}. These approaches aim to identify meaningful temporal structures that are informative for classification. While effective at capturing localized patterns, they often require extensive manual parameter tuning and face scalability challenges in high-dimensional settings.

The third approach uses \textbf{explainable AI techniques} to improve the interpretability of deep learning models, which are known for their high accuracy on time series tasks. A common method is to identify influential segments within a time series instance~\cite{siddiqui2019tsviz}. However, such instance-specific explanations are difficult to generalize across an entire dataset. To address this limitation, prototype-based approaches have been proposed~\cite{gee2019explaining}. These methods learn representative prototypes directly in the latent space and decode them back into human-interpretable forms. Despite their promise, two interpretability challenges remain: (1) the prototypes often represent ``averaged'' or compounded features over the entire time series, limiting their specificity; and (2) the decoded prototypes are not always inherently self-explanatory and may appear noisy or ambiguous.


Among the approaches discussed above, shapelets (the second category) offer one of the most direct forms of interpretability by capturing short subsequences of time series that are highly discriminative (e.g., a sharp drop followed by a rise in an ECG signal). These patterns may be directly aligned with domain knowledge, allowing experts (e.g., clinicians) to visually compare how well a shapelet matches a segment of a new time series. However, shapelets have important limitations: they do not capture global temporal structure, are sensitive to temporal misalignment and warping, and are difficult to integrate into end-to-end deep learning models due to non-differentiability. Furthermore, shapelet discovery is computationally expensive and not scalable to large datasets.

In this work, we introduce a novel joint-learning framework that addresses these challenges by reformulating shapelet discovery as a learning and optimization problem~\cite{grabocka2014learning}. Our method innovatively transforms a time series into two complementary representations: shapelet-based features that capture localized temporal patterns, and engineered features that summarize global statistical properties. By applying moving windows and feature fusion techniques, we obtain a unified representation that retains both local discriminative structure and global context. These fused features can then be seamlessly integrated into state-of-the-art deep learning models (e.g., Transformer\cite{vaswani2017attention}), for end-to-end learning with interpretable components.

\subsection{Time Series Classification}

Time series classification refers to the task of predicting a class label for an unlabeled time series by learning a mapping from training instances to their labels.
It has broad applications in healthcare~\cite{bock2021machine}, finance~\cite{sezer2020financial}, speech and activity recognition~\cite{gupta2020fault}, among others~\cite{mahmud2023review}. Public repositories such as UCR~\cite{bagnall2018uea} and UCI~\cite{asuncion2007uci} have accelerated the development and benchmarking of a wide range of these methods.

Existing methods can be grouped into five categories. 
\textbf{Similarity-based methods}~\cite{lucas2019proximity} compare time series using distance metrics (e.g., Dynamic Time Warping) and classify new examples based on their similarity to known instances. 
Dictionary-based methods~\cite{glenis2022scale} convert time series into symbolic representations and classify based on pattern frequency.
\textbf{Feature-based methods}~\cite{lines2018time} extract descriptive statistics or domain-specific features from time series intervals to form a fixed-length vector for classification. 
\textbf{Shapelet-based methods}~\cite{hills2014classification,karlsson2016generalized,le2024shapeformer,liu2025learning} identify discriminative subsequences and transform time series into feature vectors based on shapelet distances.
\textbf{Deep learning-based methods}~\cite{mohammadi2024deep} automatically learn complex features from raw time series using deep neural networks.
Finally, \textbf{ensemble-based methods}~\cite{shifaz2020tschief} combine multiple classifiers from different categories to improve overall performance through diversity.

Each method offers distinct advantages and is suitable for different types of time series and classification scenarios. Inspired by these categories, we introduce a hybrid structure that combines the interpretability of shapelet-based representations with the modeling capacity of deep neural networks. Specifically, we convert the shapelet discovery process into a differentiable learning task and aim to identify a set of representative shapelets that form a distance-based feature vector. This vector is fused with statistical features to form a rich representation of the input. We then train a deep classifier---based on a Transformer model---on this fused representation. The underlying intuition is that higher classification accuracy indicates that the learned shapelets are indeed representative and discriminative. Our approach strikes a practical balance between interpretability, accuracy, and scalability.

\subsection{Visualizing Time Series Data}
Time series visualization plays a critical role in supporting exploratory data analysis and decision-making. A wide variety of techniques exist for visualizing and structuring time series, each differing in how the timeline is encoded and in their suitability for specific tasks~\cite{aigner2011visualization,brehmer2016timelines}. Standard representations, such as line and area charts, are commonly used for their simplicity and direct mapping of temporal trends.

However, alternative designs have been proposed to address specific analytical needs. For instance, spiral-shaped axes~\cite{weber2001visualizing} are effective for highlighting cyclic patterns, while calendar layouts~\cite{van1999cluster} emphasize individual dates to reveal patterns within discrete time intervals.
Multivariate time series data, which are particularly challenging to interpret, can benefit from superposed line graphs, braided graphs, or small multiples~\cite{javed2010graphical}.
Additionally, glyph-based designs have gained attention for their compactness and expressiveness~\cite{fuchs2013evaluation}, and multiple coordinated views are often employed to provide complementary perspectives~\cite{liu2022mtv,Chen_PacificVis2020}.

Building on these foundational visualization approaches, more advanced visual analytics systems have been developed to specifically support time series clustering~\cite{sacha2017somflow, ruta2019sax}, classification~\cite{walker2015timeclassifier}, and the visualization of temporal patterns~\cite{meschenmoser2020multisegva, yu2022pseudo}. \sidecomment{C1}\revision{In which, visualization for repeating time series patterns~\cite{li2012visualizing}, is one of the closest existing works that matches our needs. 

Taking inspiration from existing work, we introduce a hybrid approach to visualizing and interacting with time series data. We develop a visual analytics system for signature-based time series exploration. The system combines state-of-the-art visualization techniques with interaction mechanisms tailored to support a streamlined and effective human-in-the-loop exploration of discriminative time series patterns.
}
\newcounter{descriptcount}
\newlist{deflist}{description}{1}
\setlist[deflist,1]{%
  before={\setcounter{descriptcount}{0}%
          \renewcommand*\thedescriptcount{\arabic{descriptcount}}},
        font={\bfseries\stepcounter{descriptcount}Definition \thedescriptcount~}
}
\section{Learning Time Series Signatures}\label{sec:method}
This section presents our method for learning discriminative temporal patterns, which we refer to as \textit{time series signatures}. We begin by formally defining the problem and associated notation. We then describe the representation based on localized temporal subsequences (shapelets) and introduce a joint learning framework that integrates both structural and statistical features.

\begin{figure*}[!t]
    \centering
    \includegraphics[width=\linewidth]{alg_workflow-v2.pdf}
    \caption{Workflow of backend algorithm and data processing. For the Preprocessing stage (A), the time-series is filtered, segmented, and normalized into a determined length and range. Then, we perform Signature Initialization (B), leveraging perceptual shapelet extraction~\cite{le2022learning}. Lastly, we train a transformer model (C) in an end-to-end manner with joint feature training to iteratively learn shapelet values.
    }
    \label{fig:workflow}
\end{figure*}

\subsection{Problem Formulation}
We define a multivariate, binary-class time series dataset as a three-dimensional tensor: 
\[
X \in \mathbb{R}^{N \times D \times m},
\]
where $N$ is the number of samples, $D$ is the number of dimensions (variables) per time step, and $m$ is the number of time steps. The corresponding class labels are $y \in \{0, 1\}^N$.
Each time series instance is denoted as:
\[
T = \{t_1, t_2, \dots, t_m\} \in \mathbb{R}^{D \times m},
\]
Our objective is to identify a set of class-discriminative subsequences:
\[
\mathcal{K} = \{S_1, S_2, \dots, S_k\}, \quad S_j \in \mathbb{R}^{D \times l_j}, \quad l_j < m,
\]
where each $S_j$ is a signature of variable length $l_j$ intended to capture localized, class-specific temporal structure.
\begin{deflist}
    \item \textbf{Signature-to-Series Distance}.\label{def:distance}
    Given a signature $S \in \mathbb{R}^{D \times l}$ and a time series $T \in \mathbb{R}^{D \times m}$, we define the distance between $S$ and $T$ as:
    \[
    \text{SeqDist}(S, T) = \min_{s' \in \mathcal{S}(T, l)} \text{Dist}(S, s'),
    \]
    where $\mathcal{S}(T, l)$ denotes the set of all contiguous subsequences of $T$ of length $l$, and $\text{Dist}(\cdot,\cdot)$ is a distance function. Unless stated otherwise, we use Euclidean distance.

    \item \textbf{Information Gain}.\label{def:information-gain}
    To assess the discriminative quality of a signature, we use a threshold-based partition. Given a signature $S$, we compute its distance to all samples and select a threshold $\theta$ to split the dataset into two subsets $X_1$ and $X_2$.
    The information gain is then defined as:
    \[
    \text{IG} = I(X) - \left( \frac{|X_1|}{|X|} I(X_1) + \frac{|X_2|}{|X|} I(X_2) \right),
    \]
    where entropy $I(X)$ is computed as:
    \[
    I(X) = - \frac{|A|}{|X|} \log \left( \frac{|A|}{|X|} \right) - \frac{|B|}{|X|} \log \left( \frac{|B|}{|X|} \right),
    \]
    with $|A|$ and $|B|$ denoting the number of class-0 and class-1 samples in $X$, respectively.

    \item \textbf{Signature-Based Representation}.\label{def:sig-representation}
    Given a set of identified signatures $\mathcal{K}$, we apply shapelet transformation~\cite{hills2014classification} to derive a feature matrix:
    \[
    M_{\text{sig}} \in \mathbb{R}^{N \times k}, \quad
    M_{\text{sig}}^{ij} = \text{SeqDist}(T_i, S_j),
    \]
    where each element represents the distance between a sample $T_i$ and a signature $S_j$.
\end{deflist}

\noindent
While our learning framework (Section~\ref{sec:learning-framework}) generalizes to multivariate and multi-class settings, our current visualization system (Section~\ref{sec:visual}) is designed for univariate, binary-class analysis. For clarity, we focus on the univariate case ($D = 1$) in the remainder of this paper.

\subsection{Joint Learning Framework}
\label{sec:learning-framework}
\vspace{3pt}
\noindent\textbf{From Exhaustive Search to Learning.}
Traditional shapelet discovery methods (e.g.,~\cite{ye2009time}) rely on exhaustive search to identify the single most informative shapelet that maximizes information gain. This process, rooted in decision tree induction, is computationally intensive, as it evaluates all possible subsequences across all series. To reduce search complexity, several heuristics have been proposed, such as random sampling~\cite{renard2015random}. However, these heuristics reduce the expected quality of discovered shapelets due to restricted search spaces. \sidecomment{C1}\revision{Such challenges prevent shapelet-based methods from becoming more widely adopted in practice.}

A more efficient alternative was introduced by Grabocka et al.~\cite{grabocka2014learning}, which reformulates shapelet discovery as a differentiable optimization problem. In this approach, shapelets are treated as trainable parameters updated via backpropagation. This formulation links shapelet quality directly to classification performance but requires the number and lengths of shapelets to be predefined, which limits flexibility and adaptability across different domains.


\vspace{3pt}

\noindent\textbf{Signature Initialization via PPSN.}
To address these computational limitations, we adopt the Perceptual Position-aware Shapelet Network (PPSN)~\cite{le2022learning, le2024shapeformer}. 

PPSN identifies perceptually salient regions by sampling around key points in each time series and extracting variable-length segments. This effectively reduces the time cost by avoiding checking every time point on each series. It utilizes the Perceptual Important Points (PIPs) method by recursively finding the index \textit{T} with the highest Perpendicular Distance (PD). PD between one point $pos$ and PIPs is defined as

\[
PD(pos, PIPs)=\frac{a\times P_{pos}-T_{pos}+c}{\sqrt{a^2+1}},
\]
where $a=\frac{T_e-T_s}{P_e-P_s}, c = T_e-a\times P_e$ and $P = z\text{-}norm([1, ..., n])$ is a normalized list of position. $T_s, T_e$ represent the value of the starting and ending points in time series $T$; similarly, $P_s, P_e$ represent the starting and ending points of the normalized list. Thus, candidates are generated by these perceptual important points, i.e. they are segments with starting and ending points set by above method. Ordered by their information gain, the top $k$ segments are used as initial signatures.
This initialization reduces search cost, promotes pattern diversity, and has been shown to produce high-quality candidates that accelerate convergence in joint learning.

\vspace{3pt}
\noindent\textbf{End-to-End Joint Learning Model.}
\revision{On top of this application, we extend the PPSN-initialized shapelets with a joint learning framework that refines them through gradient-based optimization.} The process involves:
\begin{itemize}[noitemsep, topsep=0pt]
    \item A \textit{signature transformation operator} that computes distances between sliding windows and a set of learnable signatures,
    \item A \textit{joint feature fusion module} that combines structural and statistical features at each window position,
    \item A \textit{Transformer encoder} that captures temporal dependencies across the windowed inputs,
    \item A \textit{classification head} that predicts output labels.
\end{itemize}
Each signature $S_j$ is a learnable parameter with variable length $l_j$. \revision{
 \sidecomment{C1}Unlike conventional shapelet-based pipelines that summarize a time series into a single global feature vector—where local patterns can be easily overlooked—we adopt a window-based signature representation. A fixed-length sliding window is applied across each sample, and signature distances are computed for every window. This design is particularly important for long-sequence analysis: the segmentation operation in preprocessing often yields shorter, mutually unaligned subsequences, and a single global representation fails to preserve \emph{where} and \emph{how often} a signature appears. In contrast, the window-based representation captures recurring or evolving local patterns, improves robustness to temporal shifts and misalignment, and produces a structured sequence of feature vectors that is well suited for Transformer-based modeling. Our method results in a feature tensor:
$
Z_{\text{sig}} \in \mathbb{R}^{N \times k \times w},
$
where $w$ is the number of sliding windows per sample. The $q$-th slice, $Z_{i,:,q}$, encodes the signature-based features (refer to Definition 3) for window $q$ of sample $T_i$.

While signature features provide interpretable structural patterns, they may omit broader statistical properties that are also informative for classification. To enhance representational power, we extract $r$ statistical features (e.g., median, root mean square, wavelet standard deviation) from each window. These are assembled into:
$
Z_{\text{stat}} \in \mathbb{R}^{N \times r \times w}.
$
We concatenate the signature and statistical features for each window to obtain a fused representation:
$
Z_{\text{joint}} \in \mathbb{R}^{N \times (k + r) \times w}.
$
This joint feature sequence is passed to a standard Transformer encoder (described in the supplementary material), which models temporal dependencies across windows. The output is fed into a classification head to produce predicted labels \( \hat{y}_i \) for each input \( T_i \).}
The model is trained using the binary cross-entropy loss:
\[
\mathcal{L}_{\text{BCE}} = - \frac{1}{N} \sum_{i=1}^{N} \left[ y_i \log(\hat{y}_i) + (1 - y_i) \log(1 - \hat{y}_i) \right],
\]
where \( y_i \in \{0, 1\} \) is the true class label.
Gradients are propagated through the Transformer, classification head, and signature transformation operator, allowing both the classifier and the signature parameters to be jointly optimized. 

\vspace{3pt}
\noindent\textbf{Advantages.}
This joint learning framework integrates structural and statistical information over sliding windows, supports variable-length signatures, and captures both recurring and evolving patterns in long sequences. Compared to earlier methods that decouple shapelet discovery and classifier training~\cite{ye2009time, renard2015random}, our framework unifies them into a single, differentiable process. 
This unified architecture preserves the interpretability and class-discriminative power of the learned signatures while directly enhancing model performance. Moreover, because the signatures are refined jointly with the classifier, redundant or non-informative signatures naturally receive weaker gradient updates, whereas discriminative ones are reinforced. This implicit learning dynamic guides the final signature set toward diverse and class-relevant patterns.

\subsection{System Workflow}
The overall system consists of three phases (Fig.~\ref{fig:workflow}): 
(1) data preprocessing, 
(2) signature initialization, and
(3) model training. 

\vspace{3pt}
\noindent\textbf{Phase 1: Data Preprocessing.}  
Raw time series are first cleaned and standardized through:
\begin{itemize}[noitemsep, topsep=0pt]
  \item \textit{Filtering:} Removal of corrupted or uninformative sequences;
  \item \textit{Segmentation:} Dividing sequences into uniform-length segments for training;
  \item \textit{Normalization:} Scaling values to $[0,1]$ to avoid magnitude bias.
\end{itemize}

\vspace{3pt}
\noindent\textbf{Phase 2: Signature Initialization.}  
Using PPSN, we identify salient breakpoints and extract local segments. The top $k$ segments (ranked by information gain) form the initial signature candidates.

\vspace{3pt}
\noindent\textbf{Phase 3: Model Training.}  
Each time series is processed through a sliding window, generating both:
\begin{itemize}[noitemsep, topsep=0pt]
  \item A signature feature tensor $Z_{\text{shapelet}} \in \mathbb{R}^{N \times k \times w}$;
  \item A statistical feature tensor $Z_{\text{stat}} \in \mathbb{R}^{N \times r \times w}$.
\end{itemize}
These are concatenated into the joint tensor $Z_{\text{joint}} \in \mathbb{R}^{N \times (k + r) \times w}$, at each time step, and passed to the single Transformer. The model learns to classify and simultaneously refine the signature parameters.

\vspace{3pt}
\noindent
After model training, we obtain a finalized set of $k$ signatures, denoted $\mathcal{K}$, along with additional outputs such as pairwise distances between signatures and samples, and the information gain for each signature. These outputs are passed to the visual analytics system (Section~\ref{sec:visual}) for interactive exploration.
\newlist{tasklist}{description}{1}
\newlist{reqlist}{description}{1}
\setlist[tasklist,1]{%
  before={\setcounter{descriptcount}{0}%
          \renewcommand*\thedescriptcount{\arabic{descriptcount}}},
        font={\bfseries\stepcounter{descriptcount}\textit{T.\thedescriptcount~}}
}
\setlist[reqlist,1]{%
  before={\setcounter{descriptcount}{0}%
          \renewcommand*\thedescriptcount{\arabic{descriptcount}}},
        font={\bfseries\stepcounter{descriptcount}\textit{R.\thedescriptcount~}}
}
\section{Requirement Analysis}\label{sec:req}

In this section, we first describe the data collection and how we deal and clean the datasets. Then, we summarize the tasks identified from the interviews with two domain experts who offered the medical dataset in our project. Finally, we collect the design requirements of our system.

\subsection{Task Analysis}
This research problem was inspired and characterized through the collaboration with domain experts from the health center of our university. They have been collecting birth related data since 2014, and they have made a lot of efforts to predict the preterm birthing pattern from those sensoring signals. However, the raw collected data is noisy and contains a lot of redundant information. Initially, they have tried on feature-based machine learning method which provides an acceptable result. However, the interpretability of such method is limited. The statistical features are too abstract for domain experts to intuitively understand.
\begin{tasklist}
    \item What are the major parts of an input signal that affect the labeling?\\
    When observing an individual time-series, it is difficult to efficiently describe how the trend of the data value along the time axis. For example, there are predefined peaks in a ECG data that share medical information, which directly describe significant parts of a cycle of heartbeat impulse. The interval between such regions and data range of those peaks is some key points to differentiate cardiological diseases. However, most of the condition, such predefined regions and aligned signals do not exist. Moreover, it is possible that most parts of a series have low information or even redundant. There is a need to extract intuitive and informative local shape features individually. 
    
    \item What is the overall behavior of an input dataset?\\
    Globally, what are the features that affect the overall behaviors? Existing traditional methods are able to provide summarizing information of each sample; however, the full picture of the dataset remains unknown. Through the discussion with domain experts, several  patterns that summarizes the dataset would be helpful. This also leads to a novel time-series feature representation that is completely not based on any existing works. The task for pattern discovery does not only contain local information but should have a global meaning. 
    
    \item How does the local pattern look like?\\
    Visualization is the most intuitive and interpretable way to unveil the information hidden in a complicated large dataset, especially when we directly extract and produce sequential patterns. Most of the domain experts suggest that shape or sequential features are much more intuitive than any pure numerical features. However, there is no existing system that specifically aims to achieve this goal. Although there is some event sequence visualizations, they are mostly focusing on occurrence of predefined events ~\cite{scimone2023marjorie, bernard2024ivesa} but not how the event sequence's trend and meaning. This inspires us to think of temporal local patterns and the visual representation. 
\end{tasklist}

\subsection{Design Requirements}

\begin{reqlist}
    \item Overview and Multiscale exploration\\
    The exploration always starts with an overview, offering a high-level view and sense of the data, allowing the users to understand the general trend at their first glance. As user explore firther, they may identify some critical points that prompt them to dive deeper into the data. This leads to the introduction of a scale-adjustable function, enabling users to zoom in on specific area for detailed analysis. For scaling feature, the function to simultaneously control different parts of the system is required, since it allows the correlated detailed observation. 
    
    \item Interpretation in Global-Level and Individual-Level\\
    For each feature, an global-level interpreation and an individual one are required. The global-level interpretation should provide a comprehensive overview of the feature’s behavior across the entire dataset, identifying the patterns and  correlations. On the other hand, the individual-level interpretation should focus on how the feature impacts each specific instance or data point, offering insights into its significance in the context of individual cases. Together, these interpretations will help in understanding both the overall distribution and the nuanced behavior of the feature across different instances. This requirement perfectly aligned with the characteristics of a 
    
    \item Intuitive Visual Design\\
    Aiming for a visual analytics system that provides interpretation or explanation, it is required that the visual design should be intuitive. The concept of signature is relatively abstract for non-data-analysts, including how to extract and how the classifier training affects the results. In addition, familiar visual encoding can enhance understanding and reduce the cognitive efforts to interpret the representation. Due to the above consideration, we are not aiming a super novel design. Rather than that, an appropriate arrangement of familiar visualization views that illustrate the how the features look is the most suitable direction. Thus, an intuitive visual design with the above properties supports the visual analytics.
    
\end{reqlist}


\section{\system}
We introduce \system, a visual analytics system designed to support the exploration and interpretation of discriminative time series signatures, building on the joint learning framework described in Section~\ref{sec:learning-framework}. \system~integrates model outputs, interactive visualization, and user-driven analysis to enable signature-centered investigation of binary-class time series data. It supports both global and instance-level perspectives, allowing users to examine data patterns, interpret learned signatures, and refine exploratory insights.

\subsection{Design Requirements}\label{sec:req}
We have collaborated with two domain experts from the university's health center: a professor specializing in maternal health, a PhD researcher focused on ECG-based heart disease prediction, and a data engineer with expertise in physiological and clinical data processing. The professor brings domain knowledge in perinatology, particularly in using ECG and contraction signals to support obstetric care. The PhD researcher focuses on machine learning for cardiovascular risk assessment, while the data engineer provides guidance on handling noisy, incomplete, and heterogeneous medical time series. Our collaboration involves two datasets: the first is an ECG dataset, where each time series represents a single cardiac cycle labeled as normal or abnormal; the second is a uterine contraction dataset collected via external tocometry, with each sample labeled according to whether the patient later experienced preterm birth.

While \system~is designed to be domain-agnostic, this collaboration has been essential in identifying core analytical tasks and motivating the system design with real-world needs. Expert feedback also complements general design guidelines from prior work on time series visualization. The design requirements below reflect these insights and guide the development of \system's visual and interactive components.

\begin{itemize}[left=0pt, itemsep=-2pt, labelsep=1em, align=left]
\item[\req{R1}] \textbf{Enable high-level overview and comparative analysis.}
The system should support global inspection of the dataset to reveal overall distributions, class separability, and high-level activation of learned signatures. It should enable comparison across user-defined groups (e.g., class or cluster) to identify broad trends and variations.

\item[\req{R2}] \textbf{Support exploration of data partitions formed by signatures.}
The system should allow the dataset to be grouped based on similarity to extracted signatures. Users should be able to adjust parameters that control grouping granularity and interactively explore how different subsets of data relate to specific signatures.

\item[\req{R3}] \textbf{Provide multi-resolution views of sample–signature relevance.}
Users should be able to assess the relevance of each signature across the dataset at both the aggregate and individual sample levels. The system should summarize how groups of samples relate to each pattern while also supporting detailed comparisons for selected subsets or samples.

\item[\req{R4}] \textbf{Visualize signature alignment within original time series.}
To support interpretability, the system should display where and how extracted signatures align within raw time series. Pattern matches should be shown in context (in-situ), preserving local signal structure to help users judge the semantic and domain relevance of each match.

\item[\req{R5}] \textbf{Support rich interactions for targeted analysis.}
The system should provide interactive controls for filtering and sorting based on relevance scores, class labels, cluster assignments, or other user-defined criteria. Users should be able to reorder items (e.g., samples or signatures) to facilitate comparison and examine variation within or across groups.

\item[\req{R6}] \textbf{Support visual configuration for task-specific exploration.}
The system should provide options to adjust visual representation based on the nature of the data and analysis goals. This includes control over visual encoding (e.g., curve style, line opacity, color precedence) to better suit the characteristics of the dataset and facilitate meaningful pattern interpretation.

\end{itemize}

\subsection{Visual Design}\label{sec:visual}

\begin{figure*}[!t]
    \centering
    \includegraphics[width=\linewidth]{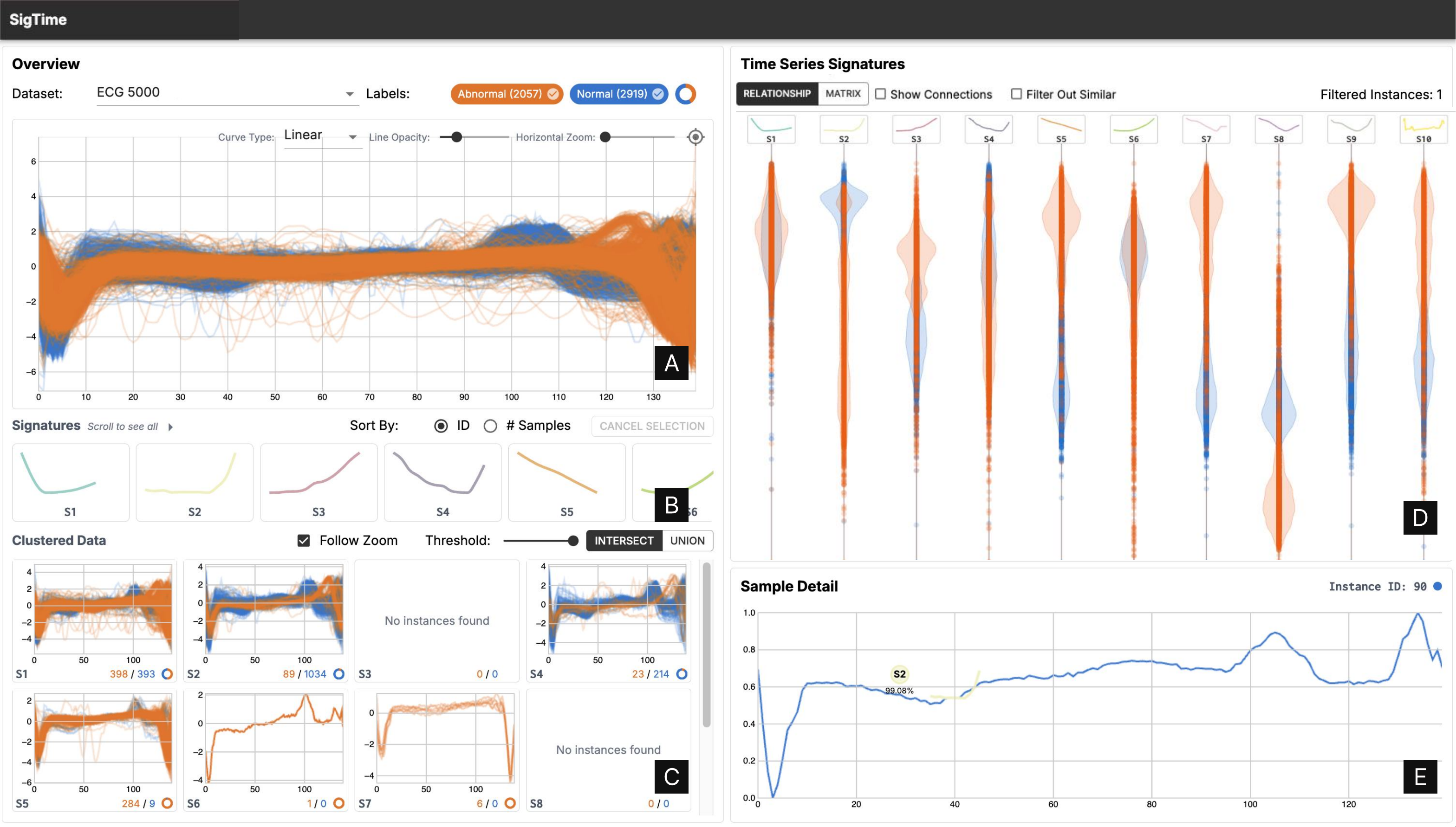}
    \caption{The interface of \system~applied to the \textit{ECG5000} dataset. It consists of three coordinated panels: the Overview Panel displays the raw time series (A), the list of learned signatures (B), and signature-based clustering of time series (C); the Signature Panel includes two switchable tabs (relationship and matrix views) for analyzing sample–signature relevance at multiple resolutions. 
    There are two checkboxes, one is to show connected lines and one is to filter out similar shapes
    (D); and the Sample Panel shows in-situ alignment of selected signatures within a chosen time series for detailed inspection (E).}
    \label{fig:teaser}
\end{figure*}

The interface (Fig.~\ref{fig:teaser}) is organized into three main panels: the \textit{Overview Panel}, which includes the raw time series, the signature list view, and the signature-based clustering view; the \textit{Signature Panel}, which provides two switchable tabs for exploring signature-to-sample relationships (relationship view) and detailed sample-level comparisons (matrix view); and the \textit{Sample Panel}, which offers a single view for in situ exploration of how signatures align with a selected sample from the matrix view.  
The exploration workflow follows a coarse-to-fine strategy, beginning with a global overview of the dataset and progressively narrowing the focus to individual samples for detailed inspection.

\subsubsection{Overview Panel}
\label{sec:overview-panel}

\noindent\textbf{Overview.}
The top-left view (Fig.~\ref{fig:teaser}-A) displays a chart of the entire time series dataset, with individual curves color-coded by class labels (e.g., blue and orange). This view provides a summary of the overall distribution patterns across classes \req{R1}. Two interactive sliders allow users to adjust the opacity of the lines and the horizontal zoom level for closer inspection \req{R6}.
Users can reduce line opacity to mitigate visual clutter when many series are overlaid in the view.
A vertical cursor appears on hover to indicate the temporal position under focus.

\textit{Curve Style Manipulations.}
The chart supports three curve styles \req{R6}: linear (left), step (middle), and basis spline curve (right). Different applications may favor different styles. For example, in ECG data, smooth curves such as
\begin{wrapfigure}{r}{0.28\textwidth}
  \vspace{-1.6em}
  \begin{center}
    \includegraphics[width=0.28\textwidth]{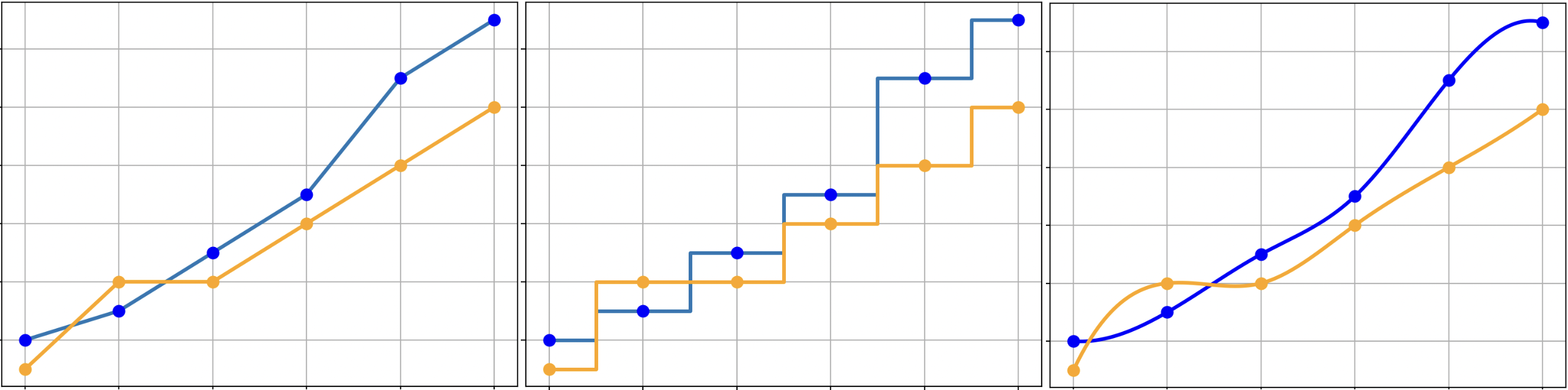}
  \end{center}
  \vspace{-1.6em}
\end{wrapfigure}
 basis splines are often preferred to highlight waveform morphology. In contrast, step charts are commonly used in discrete state transitions, such as system logs or process control data. Supporting multiple styles provides flexibility in examining the data and understanding the class-specific time series structures. 

\vspace{3pt}
\noindent\textbf{Signature List View.} 
The second view (Fig.~\ref{fig:teaser}-B) displays a ranked list of extracted time-series signatures. \begin{wrapfigure}{r}{0.15\textwidth}
  \vspace{-1.5em}
  \begin{center}
    \includegraphics[width=0.15\textwidth]{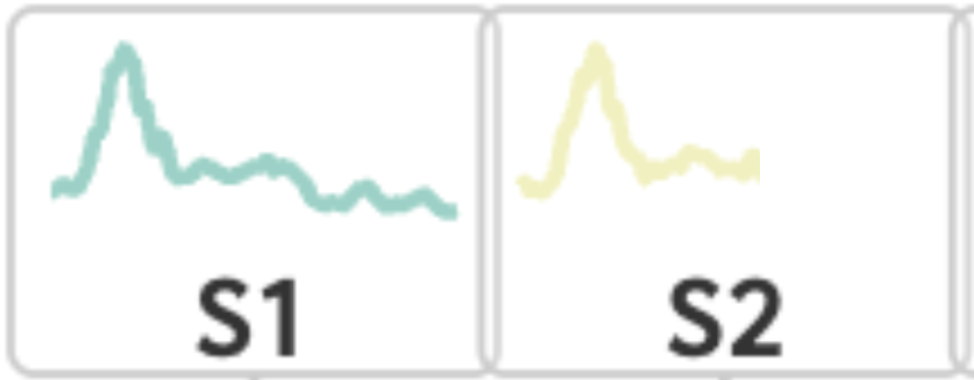}
  \end{center}
  \vspace{-1.5em}
\end{wrapfigure}
These signatures are computed in the backend (Section~\ref{sec:learning-framework}) and represent discriminative patterns identified by the classification model. Each signature is assigned a distinct categorical color. Since signature lengths may vary, they are displayed in their original form without rescaling or truncation. This design allows users to observe the shape and temporal extent of each pattern directly, providing a structural overview of the extracted signatures \req{R1}.

\textit{Ranking and Ordering.}
The list is horizontally scrollable to accommodate the full set of extracted signatures. By default, signatures are ranked by their importance scores, computed using Information Gain (refer to Definition~2) and Shapley value. The interface displays the top 10 most important signatures for exploration, although the classification model is typically trained with a slightly larger set. Users may also re-order the list based on the number of samples that match each signature \req{R5}, subject to a user-defined matching score threshold, defined as:
\[
Score_{match} = 1 - \text{SeqDist}_{\text{norm}}(T_i, S_j)
\]
where $T_i$ is a sample and $S_j$ is a signature. A higher score indicates greater similarity between the sample and the signature.

\vspace{3pt}
\noindent\textbf{Signature-based Clustering View.}  
The third view (Fig.~\ref{fig:teaser}-C) groups samples into clusters based on their $Score_{match}$ values with respect to each signature \req{R2}. Samples assigned to the same cluster exhibit similar temporal patterns, as they meet or exceed a user-defined threshold for the corresponding signature.
A threshold slider allows users to adjust the minimum $Score_{match}$ required for a sample to be included in a cluster.  
Clustering by individual signatures helps decompose the dataset into interpretable groups, each characterized by a dominant temporal pattern. This design supports targeted exploration by enabling users to examine how different parts of the data express distinct signatures.

\textit{Selecting and Filtering.}  
This clustering view also serves as an interactive filter, dynamically updating other views based on the current selection and threshold setting \req{R5}.  
When a cluster is selected (e.g., the \textit{s1} cluster), samples with $Score_{match}$ values below the threshold for that signature are excluded from the overview.  
Users can apply filters based on the \textit{union} or \textit{intersection} of multiple clusters, and adjust the matching threshold using the slider located at the top right of the clustering view. The same selection and filtering operations can also be performed from the signature list view, providing a consistent interaction model.
These interactions allow users to isolate and compare subsets of the raw time series that express specific signatures, either individually or in combination \req{R3}. This facilitates the analysis of signature occurrence patterns across the dataset and supports the identification of class-dependent or shared behaviors.

\subsubsection{Signature Panel}

\vspace{3pt}
\noindent\textbf{Relationship View.}  
The first tab presents the relationship view (Fig.~\ref{fig:teaser}-D), which aims to reveal the overall associations between all signatures and the relevant samples \req{R3} (selected from the Overview Panel).  
Each signature is represented by a small icon displayed along the top axis. From left to right, signatures follow the same ordering used in the signature list view. To only show the diverse signatures, users can check the ``Filter Out Similar'' checkbox, and the signatures are compared by using DTW (dynamic time warping) method.
Each dot on the vertical axis represents a sample, colored by its class label (blue or orange). The vertical position of a dot encodes its $Score_{match}$ with respect to the corresponding signature; dots appearing higher on the axis indicate stronger matches between the sample and the signature.
On hover on the view, the exact $Score_{match}$ value is shown as a percentage, accompanied by a horizontal reference line for clarity.

\textit{Centered Streamgraph.}  
When the number of samples is large, 
visual clutter from overlapping dots can hinder the perception of class-dependent distribution patterns.
\begin{wrapfigure}{r}{0.08\textwidth}
  \vspace{-1.9em}
  \begin{center}
    \includegraphics[width=0.08\textwidth]{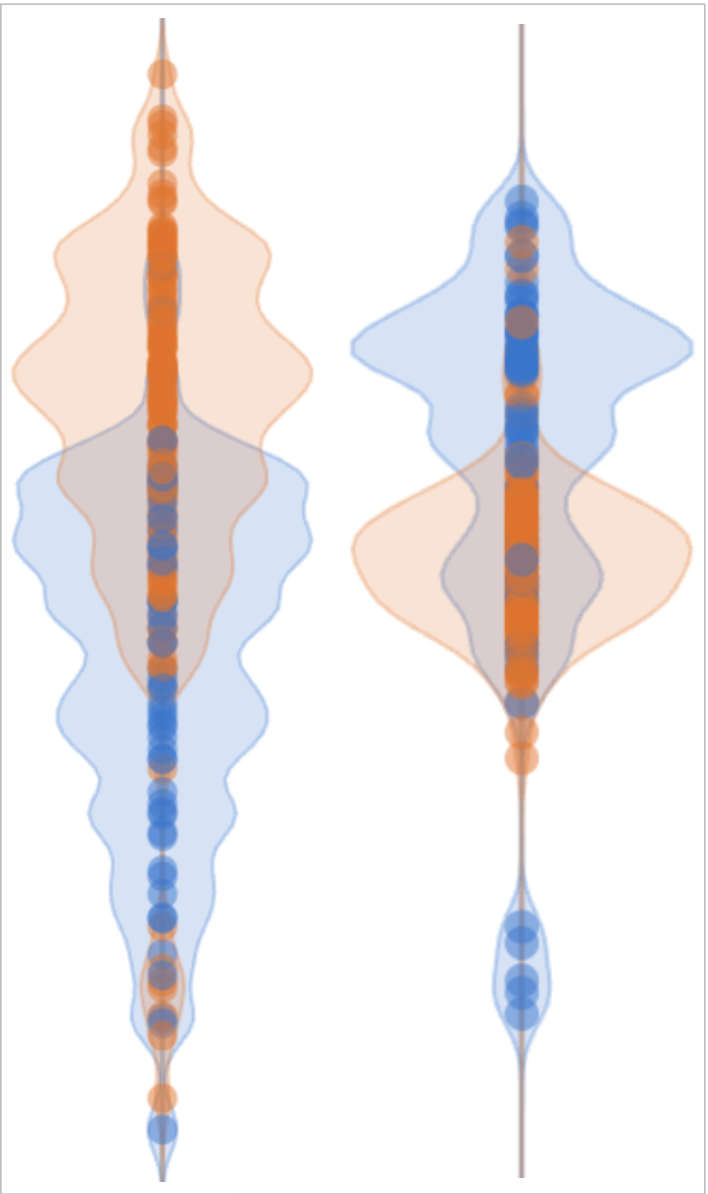}
  \end{center}
  \vspace{-1.9em}
\end{wrapfigure}
To address this issue, we introduce a centered streamgraph design. This approach is inspired by prior work~\cite{thakur20103d,krstajic2011cloudlines}, where streamgraph-like visualizations were used to depict the frequency of events over time across different locations.  
In our context, each event corresponds to a sample (dot), and instead of temporal ordering, the vertical axis represents the $Score_{match}$ between the sample and the signature.  
Each signature is treated analogously to a location in prior work, and signatures are arranged side by side to facilitate comparative analysis. To keep the same scaling for all signatures, we make the $Score_{match}$ on each signature (dimension) to be independently normalized. This makes each dimension comparable.

We apply one-dimensional Kernel Density Estimation (KDE) using a Gaussian kernel  
$G(x) = \frac{1}{\sqrt{2\pi}} e^{-x^2/2}$  
on the $Score_{match}$ distribution for each signature. To balance smoothness and local detail, we set the kernel bandwidth to $0.05$.  
The resulting density values are mapped to the width of the streamgraph, where wider regions indicate higher local sample density at a given score range.  
To ensure visual comparability across signatures, the density values are min-max normalized using the maximum density observed across all signatures.  
This design allows users to visually compare class-wise sample distributions across signatures. Large differences between class-specific densities suggest that the corresponding signature is more effective for distinguishing between classes.

\textit{Converting to Parallel Coordinate Plot.}  
While the streamgraph highlights class-level distribution differences per signature, it does not preserve the trajectory of individual samples across the set of signatures ($\mathcal{K}$).  
To enable per-sample analysis, users can toggle the ``Show Connections'' button at the top of the panel. This adds line connections across axes for all visible samples, effectively transforming the view into a parallel coordinate plot \req{R6} (Fig.~\ref{fig:preterm-workflow}-b). Line opacity is automatically adjusted based on the number of displayed samples to reduce visual clutter.
This representation supports higher-level inspection of how individual samples are characterized by the selected signatures.

\vspace{3pt}
\noindent\textbf{Matrix View.}  
The matrix view (Fig.~\ref{fig:ecg-workflow}-e) provides detailed, sample-level comparisons, where each row represents an individual sample. It displays finer-grained information about the relationship between samples and signatures \req{R3}.  
The view includes all selected samples (from the Overview Panel), and each row contains the sample ID, its assigned cluster (based on the highest $Score_{match}$), class label, and a sequence of cells corresponding to the set of extracted signatures, namely $\mathcal{K}$.  
Within each cell for a sample $T_i$, a horizontal bar encodes the $Score_{match}$ to a given signature $S_j$; longer bars indicate \mbox{stronger matches.}

\textit{Ranking and Ordering.}  
The matrix view supports four primary row ordering options \req{R5}. First, by sample ID (default); second, by cluster assignment (e.g., from $S_1$ to $S_{10}$), which helps examine how samples grouped under the same signature differ in their class labels and matching scores; third, by class label, enabling direct comparison of $Score_{match}$ distributions across classes; and fourth, by a selected signature, which ranks samples based on their similarity to that signature, facilitating focused analysis of its discriminative capacity.
These reordering interactions allow users to explore the dataset from multiple perspectives, supporting the identification of patterns both within and across clusters. They also assist in evaluating the relevance and specificity of individual signatures for distinguishing between classes.  
This view complements the relationship view by offering a more detailed, sample-level breakdown, whereas the relationship view provides a high-level overview of signature–sample associations.

\subsubsection{Sample Panel} 
\noindent\textbf{Sample Detail View.}  
The Sample Panel (Fig.~\ref{fig:teaser}-E) is designed to support in-situ exploration of how selected signatures align with a specific sample \req{R4}. Users can select a row from the matrix view to inspect its alignment.  
The design is straightforward: the complete time series of the selected sample 
\begin{wrapfigure}{r}{0.11\textwidth}
  \vspace{-1.8em}
  \begin{center}
    \includegraphics[width=0.11\textwidth]{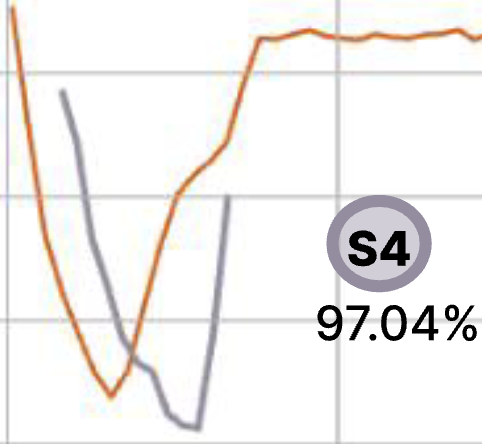}
  \end{center}
  \vspace{-1.8em}
\end{wrapfigure}
is plotted, and each signature of interest is overlaid at the location of its best-matching subsequence, identified using a sliding window approach based on minimum distance (refer to Definition~1).
For each aligned signature (e.g., $S_4$ in the right figure), the corresponding  $Score_{match}$ is displayed below each annotation to indicate the strength of the match. 
Signatures with higher scores are rendered with higher opacity to visually emphasize stronger alignments.
This view is important for validating whether a signature truly captures representative structure within a sample. 

While global views, such as the relationship or matrix view, summarize signature matching scores, they do not show where or how a signature manifests in the raw signal.  
For example, in ECG data, a signature may correspond to a particular waveform, such as a QRS complex. Visualizing the alignment in the raw time series allows users to verify whether the extracted signature correctly matches expected physiological features and whether the matching segment reflects a consistent and meaningful pattern. This form of detailed inspection is essential for interpreting the reliability and relevance of learned signatures in real-world applications.

\subsection{Coordinated Interactions}
Our system supports coordinated interactions across multiple views, including filtering and selection, synchronized zooming and panning, and global style configurations \req{R5}\req{R6}.

\vspace{3pt}
\noindent\textbf{Filtering and Selection}.
Users can apply filters based on the \textit{union} mode or \textit{intersection} mode of multiple signatures, either from the signature list view or the signature-based clustering view. The matching threshold, set as a default value according to empirical study and adjustable in the clustering view, controls which samples are considered relevant. It determines the number of samples displayed in both the clustering and signature views and influences the resulting class distribution, where only the samples with a matching score higher than the threshold are shown. This threshold also affects which signatures are visible in the detailed view, as only those exceeding the threshold are shown for each sample. 

\vspace{3pt}
\noindent\textbf{Synchronous Zooming and Panning}.  
Users can zoom and pan the raw data view via horizontal scaling or direct canvas dragging. These interactions are synchronized with the clustering view and the sample detail view, allowing close inspection of local temporal patterns and comparisons across levels of granularity. Synchronization can be toggled using the \textit{Follow Zoom} checkbox above the clustering view.

\vspace{3pt}
\noindent\textbf{Global Style Configurations}.  
To support rendering a large number of instances, users can adjust several global style settings:  
1) \textit{Curve type}, as described in Section~\ref{sec:overview-panel};  
2) \textit{Line opacity}, which modulates the visual weight of individual instances in both the raw data view and the relationship view;  
3) \textit{Render precedence}, allowing users to prioritize the drawing order of positive or negative samples to control visibility under overlap.  
To ensure responsiveness, updates to global style settings are throttled with a 500\,ms delay.

\vspace{3pt}
\noindent
In summary, the system integrates multiple coordinated views and interactions to support scalable, signature-centered exploration of class-specific and cluster-specific patterns in time series data.
\section{Example Usage Scenarios}
\label{sec:usage-scenarios}

\begin{figure*}[!htp]
    \centering
    \includegraphics[width=0.9\linewidth]{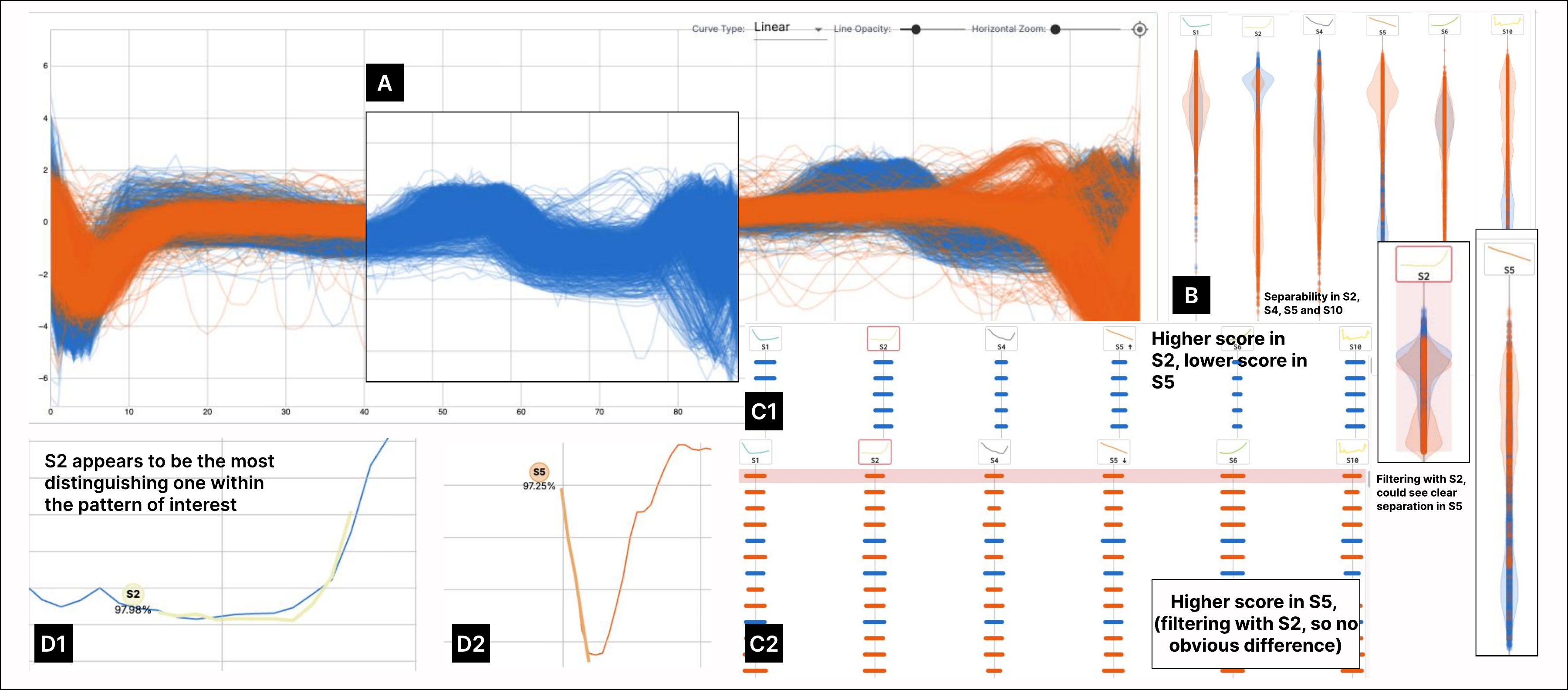}
    \caption{ The orange lines indicate abnormal samples, and the blue lines are normal samples. There is a trapezoid-like shape at the end of a normal sample (A). The signature list already excludes signatures with similar shapes. $S2$ with a flat shape is selected. On the other hand, the distribution indicates that $S2, S4, S5$ and $S10$ show the separating ability (B). The filtering with the signature $S2$ makes the separation in $S5$ much more clear. The matrix view with sorting feature helps the user quickly locate best-matched instances. These are the resulst of filtering with $S2$ and ordered by $S5$. One is in increasing order (C1), and the other one is in decreasing order (C2).One sample is selected with matching feature $S2$ (D1), while another selected sample (d2) matches with normal features $S5$ (D2).}
    \vspace{-0.1in}
    \label{fig:ecg-workflow}
\end{figure*}

We demonstrate the utility of \system~using two datasets: a publicly available ECG dataset (\textit{ECG5000}) and a proprietary clinical dataset related to preterm birth, provided by our collaborators. These datasets represent two common types of time series analysis tasks: biomedical signal classification and clinical condition monitoring. The selection was guided by domain needs for interpretable, structure-driven exploration of physiological signals. For each scenario, we describe the dataset and summarize insights gained through using \system. 
Our analyses were conducted in close collaboration with the two domain experts to ensure clinical relevance and validity of the insights.

\subsection{Scenario One: Congestive Heart Failure}

The \textit{ECG5000} dataset consists of heartbeat signals derived from a 20-hour continuous electrocardiogram (ECG) recording of a patient with severe congestive heart failure. 
\begin{wrapfigure}{r}{0.17\textwidth}
  \vspace{-1.8em}
  \begin{center}
    \includegraphics[width=0.17\textwidth]{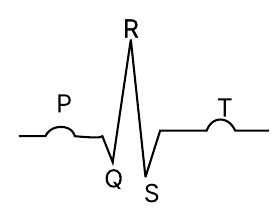}
  \end{center}
  \vspace{-1.8em}\
\end{wrapfigure}
Typically, an ECG signal can be represented as PQRST, which is composed of P wave, QRS complex, and T wave, presenting a heartbeat. Cardiovascular experts often refer to each part of such signals as peaks or wave, such as the ``S-T'' segment, to locate the abnormality that happens between the S point and the T wave. The data originates from the BIDMC Congestive Heart Failure Database (record “chf07”) available via PhysioNet~\cite{goldberger2000physiobank}. 

To prepare the dataset, individual heartbeats were segmented and resampled to a fixed length of $m=140$ time steps using linear interpolation. The original labels included five heartbeat classes—one normal and four abnormal. For binary classification, we merged all abnormal categories into a single “Abnormal” class, resulting in $N=5,000$ labeled univariate time series. Each sample was preprocessed through per-sample normalization; no additional filtering or segmentation was required beyond the original data preparation.

The experts are typically interested in identifying local waveform deviations that may indicate arrhythmic or abnormal cardiac events, such as distorted QRS complexes or premature ventricular contractions. While global classification accuracy is important, clinical interpretation requires an understanding of *where* and *how* specific heartbeat patterns differ across classes. Signatures extracted by our model provide this localized structure by highlighting temporally bounded subsequences that contribute most to class separation.

The Overview Panel revealed clear differences in signature activation between classes, suggesting that class-specific signatures were captured during training (Fig.~\ref{fig:ecg-workflow}-A). The Signature Panel showed that several signatures (e.g., $S2, S4, S5$ and $S10$ ) exhibited high separability of normal and abnormal instances, indicating their role in distinguishing pathological heartbeats. Filtering with $S2$, the other remaining signatures show even clearer distribution (Fig.~\ref{fig:ecg-workflow}-B). The Matrix View further grouped subsets of abnormal samples by dominant signature alignment, suggesting shared structural features across these signals (Fig.~\ref{fig:ecg-workflow}-C1 and Fig.~\ref{fig:ecg-workflow}-C2). In the Sample Panel, detailed inspection of signature alignment locations revealed consistent deviations in waveform morphology around known landmarks such as the QRS complex (Fig.~\ref{fig:ecg-workflow}-D1 and Fig.~\ref{fig:ecg-workflow}-D2). These alignments allowed the experts to verify whether specific waveform anomalies were systematically captured and used in classification.
This scenario illustrates how \system~ supports interpretable model analysis in biomedical signal classification. Learned signatures serve as mid-level representations that summarize recurring, class-discriminative temporal structures. Their spatial relevance and visual alignment within raw signals enable the domain experts to assess whether the model's decisions are grounded in physiologically meaningful features.

\vspace{3pt}
\noindent\textbf{Overall Difference Between Two Classes.}  
The distribution of the whole dataset demonstrates a significant difference between the two classes, presenting clear core lines for each class (\req{R1}). The normal samples appear to have a flat trapezoid-like waveform at the end of the series, which does not exist in abnormal samples (Fig.~\ref{fig:ecg-workflow}-A). This provides a rough instance-based distribution overview between classes. However, there is a scattered outlier and horizontal shifting in the view, which may come from the shifting of the segmentation of a single pulse. This may lead to relative uneasiness in detecting PQRST waves in the diseased label. ``\textit{I can see the distribution difference, but it is still vague, need to move on to further detailed views}'', said the domain expert.

\vspace{3pt}
\noindent\textbf{Labeling aligned with the clustered group.}  
The flat patterns observed in the overview view (Fig.~\ref{fig:ecg-workflow}-b) prompt the experts to further explore these patterns. The corresponding clusters (Fig.~\ref{fig:teaser}-C) also show that major members come from normal class (\req{R2}). 
Especially for signature $S2$, over $90\%$ of that cluster are from normal class. The selection of such clustered member \req{R4} leads to further exploration, focusing on the samples in cluster $S2$. 

\vspace{3pt}
\noindent\textbf{Overall relationship presenting consistent patterns.}  
First of all, referring to Fig.~\ref{fig:teaser}-D and Fig.~\ref{fig:ecg-workflow}-B, the overall relationship between all signatures presents consistent distributed patterns (\req{R3}). Specifically, for signatures $S2$ through $S10$, samples are represented as little dots forming a centered streamgraph displaying distinct distribution between the two classes. The signature $S2$ presents relatively clear patterns among all signatures, as a high concentration for normal samples. Thus, it leads the experts to filter the whole dataset with this signature. ``\textit{It is convincing that the signatures have the separating ability, which means the model is effective. However, we need to locate the signature positions to link the meaning of it}", prompting the experts to explore individual series.

\vspace{3pt}
\noindent\textbf{Samples sorted in matrix view.}  
To select each individual instance, the experts move on to the matrix view. Fig.~\ref{fig:ecg-workflow}-C1 and ~\ref{fig:ecg-workflow}-C2 show the filtered samples. They are sorted by the matching score to the signature $S5$, under the filtering with the cluster of signature $S2$, one is in increasing order, one is in the opposite order. With the filtering function (for diversity) and scaling mechanism for each signature, we can tell the instance difference by the length-encoded representation.

\vspace{3pt}
\noindent\textbf{Sample detail view revealing class-specific signature matching.} 
Subsequently, by selecting a sample with the highest matching score to signature $S2$ (\req{R5}), the sample's detailed view directly presents the matching condition (Fig.~\ref{fig:ecg-workflow}-D1). The normal sample exactly matches the flat signature $S2$, confirming the trend observed in the overview. For comparison, another sample from the abnormal class is selected (Fig.~\ref{fig:ecg-workflow}-D2), which does not match $S2$ under the defined threshold. This contrast highlights the class-specific nature of the learned signature. ``\textit{I can see it is matching to the end of the signal consistently for series coming from the same groups, but not immediately link them to actual medical meanings without the indication of PQRST positions}'', pointing out that the system is in need of a customized design to meet the requirements for each dataset.

\subsection{Scenario Two: Preterm Birth}\label{sec: preterm}

\begin{figure*}
    \centering
    \includegraphics[width=0.95\linewidth]{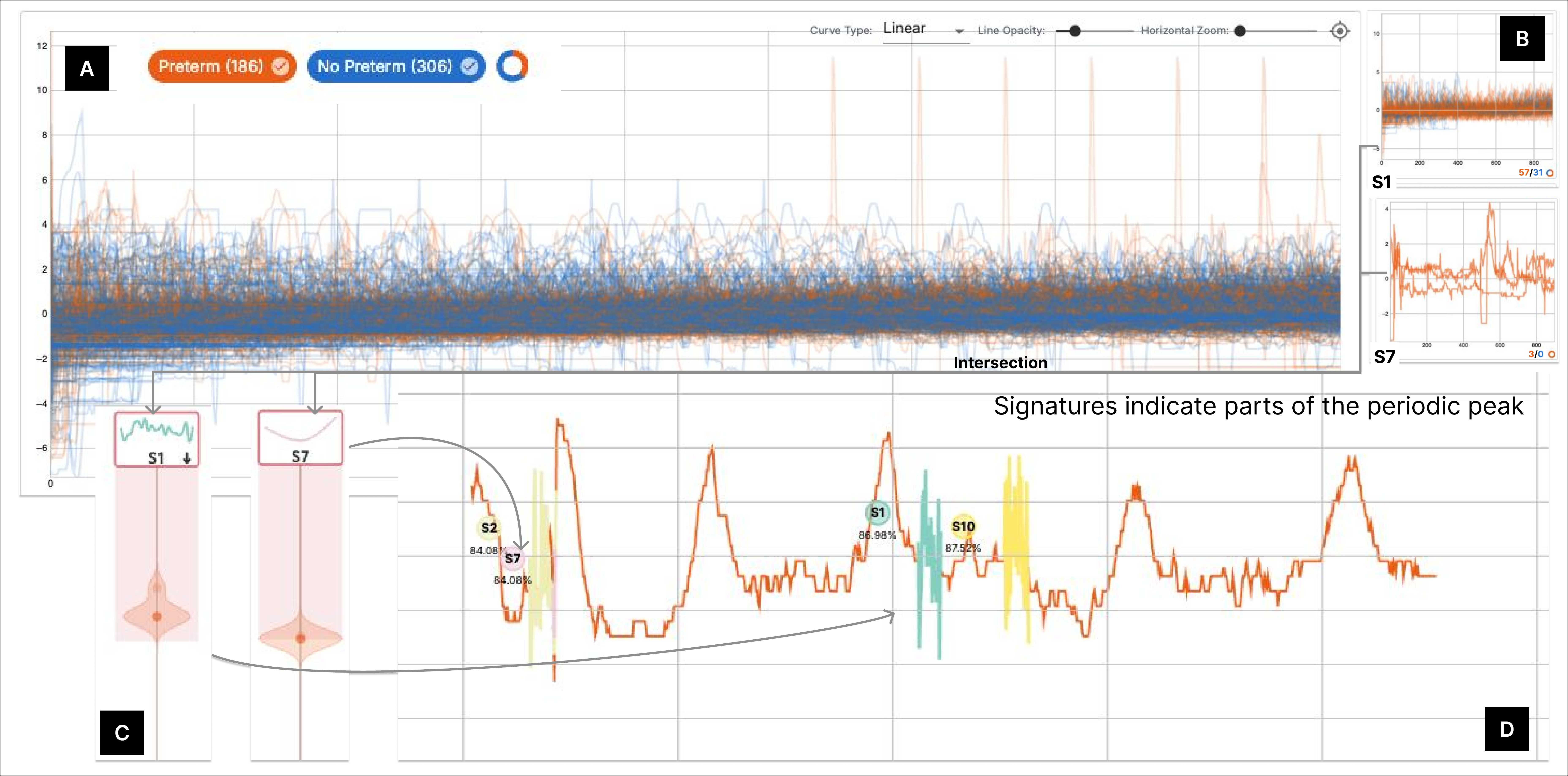}
    \sidecomment{C3}
    \caption{In the overview (A), it shows that the instances with preterm label (orange) have periodic hill-like peaks; while the ones with no preterm labeling do not have those. \revision{Select signatures $S1$ and $S7$ with majority of their clusters are preterm instances, especially $S7$ makes a perfect cut with appropriate threshold setting(B)}. The signature relationship distribution of such intersection (C) shows only instances from preterm birth class, indicating effective discriminative signatures in this class. One example (D) of such an intersection is picked from the matrix view. Two signatures are marked with respective colors, indicating the downward phase of some periodic peaks.}
    \vspace{-0.15in}
    \label{fig:preterm-workflow}
\end{figure*}

We use a proprietary dataset to demonstrate the practicality and robustness of \system~in real-world clinical monitoring tasks. The dataset was collected by our university’s Health Center as part of a preterm birth risk assessment study. Uterine contraction signals were recorded during prenatal visits to monitor whether a pregnancy would result in preterm birth. The dataset includes recordings from 80 patients, each monitored multiple times during the late stages of pregnancy. Each signal is labeled based on the eventual delivery outcome (preterm birth or full-term birth). All data were de-identified and collected under IRB-approved protocols.

Preprocessing followed the pipeline outlined in Fig.~\ref{fig:workflow}.
We first removed recordings that were too short or showed no contractions. The remaining time series were normalized to the range $(0, 1)$ and segmented using a sliding window to produce equal-length samples for model training. The resulting dataset is challenging due to its clinical noise, variability across patients, and lack of temporal alignment between samples. To prepare the dataset, individual waveforms were segmented using a sliding window of 15 minutes and resampled to a fixed length of $m=900$ time steps. The total of $n=172$ time series samples are obtained after filtering.

\vspace{3pt}
\noindent\textbf{Overview showing instance-based difference between two classes.}  
The overview (Fig.~\ref{fig:preterm-workflow}-A) presents a clutter of time-series lines, where normal and abnormal samples are mixed together (\req{R1}). Nonetheless, we can still identify the difference by the periodic peaks in preterm birth instances (orange lines). While for the no-preterm instances (blue lines), the distribution is much more irregular and cluttered. ``\textit{This may mean the upcoming birthing for those mothers as they tend to have regular urinary contractions}'', suggested by the domain experts.``\textit{But still need to move on towards other views for detailed observation.}''

\vspace{3pt}
\noindent\textbf{Signatures distribution difference between two labels.} 

 Unlike the \textit{ECG5000} dataset, it is challenging to establish a threshold for the matching score to effectively separate the dataset. \req{R2} The separation points between the two classes are unclear in the whole picture of each signature. ``\textit{It is difficult to directly recognize any pattern in this distribution}.'' Thus, we choose to switch on the filtering function.\sidecomment{C3}\revision{With appropriate setting of threshold, cluster $S7$ only contains preterm birth instances, and the cluster of $S1$ is also composed of instances mostly in preterm birth instances (Fig.~\ref{fig:preterm-workflow}-B),} we select the intersection of these two signatures (Fig.~\ref{fig:preterm-workflow}-C) to further explore the dataset. The distribution filters out no-preterm samples, and the remaining instances concentrate at a close location. Thus, we can conclude that even though the quality of signatures is not high enough to make clean patterns, these signatures still contribute to consistent distribution.

\vspace{3pt}
\noindent\textbf{Detailed view showing critical segments.}  

Sorting the instances in the intersection of signatures $S1$ and $S7$, the samples that are most effectively represented by these features are listed (\req{R4}\req{R5}). Due to the lower quality of the extracted signatures, the information provided by capturing features in an individual sample is also limited. However, this does not imply that the extracted information is redundant. Signatures represent the most discriminative patterns, and the segments extracted by sliding signatures through each series highlight the most informative and critical parts \req{R4}. In~\ref{fig:preterm-workflow}-D, several signatures show the important segments in a preterm birth instance. ``\textit{Some signatures mark the downward slope of those periodical peaks, which may be the uneasiness for unitary relaxing when monitoring preterm birth mothers.}''

 As noted in the previous paragraph, this dataset is of lower quality compared to others, making analysis more challenging. Many samples contain undesirable patterns, such as prolonged flat or step-like segments, which are difficult to remove—especially when they occur within otherwise informative regions. ``\textit{There still remains a large amount of artifacts which affects the training results; as I can still see a lot of instances are not of good quality for us to judge the results}'', said the domain expert. ``\textit{It is possible to improve by manual selection by the domain experts}'', leads to a direction of our future plans. Nevertheless, as shown above, the learned signatures can effectively focus on the meaningful portions of the signal, allowing users to bypass noisy segments and concentrate on the \mbox{discriminative patterns}.

\section{Quantitative Evaluation}
\label{sec:evaluation}

We quantitatively evaluate the effectiveness of our learning framework in generating representative and in class-discriminative signatures.

\begin{table*}[!htp]
    \centering
    \caption{Classification accuracy across datasets and models. Joint Training (our approach) achieves top or near-top performance on most datasets and demonstrates stable generalization across varying data characteristics.}
    \begin{tabular}{c|c|c|c|c|c}
    \hline
    \textbf{Dataset} & \textbf{\begin{tabular}[c]{@{}c@{}}Vanila-Transformer\\ VT\end{tabular}} & \textbf{\begin{tabular}[c]{@{}c@{}}Statistic-FE\\ S-FE\end{tabular}} & \textbf{\begin{tabular}[c]{@{}c@{}}Shapelet-FCN\\ SP-FC\end{tabular}} & \textbf{\begin{tabular}[c]{@{}c@{}}Shapelet-Transformer\\ SP-T\end{tabular}} & \textbf{\begin{tabular}[c]{@{}c@{}}Joint Training\\ JT \end{tabular}} \\ \hline
    
    Preterm               & 0.621 & \textbf{0.658} & 0.537 & 0.582 & 0.625 \\
    ECG5000-binary        & 0.970 & \textbf{0.979} & 0.960 & 0.977 & 0.970 \\ \hline
    ECG200                & 0.806 & 0.810 & 0.851 & 0.748 & \textbf{0.853} \\
    SonyAIBORobotSurface1        & 0.713 & 0.710 & 0.746 & 0.858 & \textbf{0.879} \\
    SonyAIBORobotSurface2        & 0.761 & 0.920 & 0.820 & 0.653 & \textbf{0.940} \\
    Coffee                & 0.943 & \textbf{1.000} & 0.989 & 0.664 & 0.957 \\
    GunPoint              & 0.863 & 0.923 & \textbf{0.976} & 0.832 & 0.961 \\
    BirdChicken           & \textbf{0.869} & 0.799 & 0.830 & 0.655 & 0.865 \\
    StrawBerry            & 0.910 & 0.927 & 0.911 & 0.932 & \textbf{0.950} \\ \hline
    \end{tabular}
    \vspace{-0.1in}
    \label{tab:performance}
\end{table*}

\subsection{Experimental Setup}

\noindent\textbf{Metrics.}  
While there is no direct metric to assess the effectiveness of the learned signatures, we use classification accuracy as an indirect indicator of their discriminative utility. The core assumption is that if the learned shapelets (signatures) capture class-specific temporal structures, then models using these features should perform well in standard classification tasks.

\vspace{3pt}
\noindent\textbf{Datasets.}  
In addition to the \textit{ECG5000} ($N=5000$, $m=140$) and \textit{Preterm} ($N=172$, $m=900$) datasets used in our case studies, we evaluate our method on a set of publicly available benchmark datasets from the UCR Time Series Archive~\cite{UCRArchive2018, dau2019ucr}. All selected datasets are univariate, binary-class, fixed-length, and pre-aligned. We applied only per-sample normalization without additional preprocessing.
\textit{ECG200} ($N=200$, $m=96$): Each sample represents a single heartbeat. Classes: normal vs. myocardial infarction.
\textit{SonyAIBORobotSurface} ($N=621$, $m=70$): Sensor readings from a robot navigating different surfaces, smooth or coarse.
\textit{Coffee} ($N=56$, $m=286$): Spectral measurements from coffee beans. Classes represent different varieties.
\textit{GunPoint} ($N=200$, $m=150$): Motion sequences of subjects simulating gun-draw actions.
\textit{BirdChicken} ($N=40$, $m=512$): Contour shapes extracted from bird or chicken silhouettes.
\textit{Strawberry} ($N=983$, $m=235$): Shape-based features distinguishing two strawberry cultivars.

\vspace{3pt}
\noindent\textbf{Baselines.}  
We compare our method against several representative baselines:
\textit{Vanilla-Transformer:} A Transformer encoder applied directly to the raw time series. While the Transformer may not always be the optimal architecture for time series, it is widely adopted due to its flexibility and strong performance across diverse tasks. This model serves as a reference point to gauge our methods.
\textit{Statistic-FE:} A classifier trained on engineered statistical features extracted over sliding windows. 
\textit{Shapelet-FCN}~\cite{grabocka2014learning}: A neural model that introduced the concept of learnable shapelets via differentiable optimization. It learns a small set of fixed-length subsequences.
\textit{Shapelet-Transformer:} Our proposed enhanced shapelet-based model that applies a sliding window over the input and feeds the resulting representations into a Transformer encoder. 
\textit{Joint Training:} Our final model combining statistical features and learnable shapelets within a unified Transformer-based framework. 

Lastly, the recent \textit{ShapeFormer}~\cite{le2024shapeformer} employs fixed, non-learnable shapelets and generates representations that are not directly interpretable, which differs from our objective. Therefore, we exclude it from comparison. All models are trained using the same data splits and standard hyperparameters unless otherwise specified.

\subsection{Results}

Table~\ref{tab:performance} summarizes the classification accuracy across all datasets and baseline models. We highlight two main observations:

\vspace{3pt}
\noindent\textbf{Different methods exhibit strengths on different datasets; JT achieves the most stable performance overall.}  
Statistic-FE performs well on noisy or heterogeneous datasets such as \textit{Preterm}, \textit{ECG5000}, and \textit{Coffee}, where handcrafted features may effectively summarize coarse global trends. In contrast, shapelet-based models (SP-FC, SP-T, JT) generally perform better on datasets where local discriminative structures are more prominent, such as \textit{GunPoint}, \textit{Strawberry}, and \textit{SonyAIBORobotSurface}.
Our proposed \textit{Joint Training (JT)} method consistently avoids extreme failures and performs competitively across all datasets. For instance, while SP-T and SP-FC underperform significantly on \textit{Preterm}, falling below even the VT baseline, JT remains close to the top-performing model (S-FE). This indicates improved robustness and generalization.

\vspace{3pt}
\noindent\textbf{JT achieves top or near-top accuracy on most datasets.}  
JT outperforms or matches the best-performing models across a majority of datasets. Compared to VT and S-FE, JT offers a stronger balance between capturing fine-grained local structures (via shapelets) and global context (via Transformer-based modeling with moving windows). For datasets where VT or S-FE perform best (e.g., \textit{Preterm}, \textit{BirdChicken}, \textit{Coffee}), JT remains highly competitive with only marginal differences. Notably, for several datasets such as \textit{SonyRobotSurf1\&2}, \textit{GunPoint}, and \textit{Strawberry}, JT achieves the highest accuracy overall.
These results support our claim that JT provides interpretable and discriminative representations while maintaining competitive performance across diverse time series domains.

\section{Implications, Limitations, and Future Work}

\vspace{3pt}
\noindent\textbf{General Applicability.}  
This work presents an application-agnostic system for signature-centered time series analysis, designed to answer the question: what local structures differentiate time series classes? Our learning framework emphasizes a balance between interpretability and accuracy by extracting class-representative local patterns. We visualize these patterns through intuitive and task-driven visual encodings, based on carefully derived design requirements.
The system can be extended for domain-specific tasks. In clinical settings, for example, future versions may incorporate QRS annotations or support switching between ECG leads, enhancing interpretability and contextual relevance for expert users.

\vspace{3pt}
\noindent\textbf{Learning Framework Scalability.}  
Our method demonstrates good scalability in terms of computational efficiency. On a machine with a 4.80GHz  Intel(R) Xeon(R) w7-3465X CPU, 512GB RAM, and an RTX 4090 GPU, the framework trains on ECG5000 ($N=5000$, $m=140$) and Preterm ($N=172$, $m=900$) datasets within one hour. Additional experiments on synthetic datasets with varying $N$, $m$, and hyperparameters (e.g., number of shapelets $k$, window size) are included in the appendix.
Compared to traditional shapelet methods that typically support only small datasets (often $N<200$ and $m<100$), our method scales to larger datasets efficiently, especially with GPU acceleration. This makes it practical for modern, real-world time series applications.

\vspace{3pt}
\noindent\textbf{Visualization Design Scalability.}  
Our current interface supports visualizing thousands of time series efficiently with Canvas rendering. However, browser performance may degrade with substantially larger datasets, limiting the ability to display raw series at full scale. To address potential visual clutter, techniques such as heatmap encoding can be employed in the overview. The optimal solution, however, depends on the characteristics of the dataset. For instance, the preterm birth dataset used in our second scenario does not exhibit the regular patterns typical of ECG data. In such cases, showing all raw series in the overview may introduce certain level of confusion. A more effective approach could be to group signals into representative clusters, or in some cases omit the overview entirely, to ensure the visualization remains meaningful and interpretable. Similarly, when the sequences are very long, alternative design strategies may be needed—such as hierarchical summarization or multi-resolution views—to maintain scalability and interpretability.

We acknowledge the downside of merging several classes into one label, as it may mask some critical patterns. The current system can be extended to multi-class scenarios with minimal changes, mainly involving visual encoding adjustments (e.g., color palette control for signatures and samples). In contrast, supporting multivariate time series poses more substantial challenges. While our learning framework natively supports multivariate inputs, effective visual exploration of such data remains an open research problem. For example, the alignment of signatures on different dimensions is a crucial problem that remains to be explored, as the signature was proposed and targeted univariate datasets. Future work will investigate advanced designs for interactive multivariate signature visualization.

\vspace{3pt}
\noindent\textbf{Customized Design and Human Involvement.}
Our work initially aimed to address general time series analysis problems by linking classification with feature recognition, and the results demonstrate the feasibility of an end-to-end, training-based analysis system. However, feedback from our domain experts highlighted that, although different medical datasets may share waveform-like structures, they often require dataset-specific processing to generate meaningful interpretations. For example, constraining certain signatures to specific subsequences (e.g., segments of the PQRST wave) can provide more clinically relevant insights. At the same time, incorporating active human involvement is critical. Currently, users can adjust only a limited set of parameters, without the ability to manually filter out artifacts or define custom patterns. While our method is designed for general applicability, addressing domain-specific needs---particularly in medical contexts---will require adding richer interactive capabilities in future work. \sidecomment{C2}\revision{One potential solution is to allow users to choose how the signatures are initialized. Users could specify desired patterns through query methods such as hand-drawn sketches or natural-language descriptions (e.g., \cite{shapesearch2020, hand2018}). This form of guided initialization provides model training with more meaningful starting points.

To further incorporate human sense-making into the workflow, the system should support mechanisms for users to assign semantic meaning to extracted patterns—or reject patterns that are irrelevant or undesirable. In addition, uncertainty must be appropriately addressed: extracted patterns may lead to multiple plausible interpretations. This can be mitigated by enabling users to annotate patterns or the corresponding sample-level views directly, facilitating collaborative interpretation and improving the reliability of the resulting explanations.
}

\vspace{3pt}
\noindent\textbf{Toward Real-World Decision-Making.}  
A key future direction is integrating our system into real-world decision-support workflows. While the current system identifies discriminative local patterns, the next step is to translate these insights into actionable decisions. For example, in the Preterm dataset, once high-risk waveform signatures are identified, further questions arise: How early can risk be reliably predicted? What interventions could be triggered based on signature activation? Addressing these questions requires close collaboration with domain experts to link interpretable model outputs to clinical decision protocols, an important direction for future work.

\section{Conclusion}

We present \system, a visual analytics system for learning and exploring time series signatures. At its core is a learning framework that reformulates signature (shapelet) discovery as a machine learning optimization problem. By jointly modeling structure-based (shapelet) and statistic-based representations, our method enables end-to-end training with robust, balanced performance across diverse time series classification tasks. The visual interface supports signature-centered exploration, allowing users to assess learned signatures in the context of raw data and interpret class-discriminative local patterns. We demonstrate the effectiveness of \system~through two usage scenarios—biomedical signal classification (\textit{ECG5000}) and clinical condition monitoring (\textit{Preterm})—and evaluate it on multiple benchmark datasets. Results show that our framework yields interpretable, discriminative features while maintaining competitive accuracy and scalability. A key future direction is to integrate \system~into real-world decision-support workflows, establishing a foundation for interpretable, signature-driven time series analysis across domains.

\section*{Acknowledgments}

This research is supported in part by the National Science Foundation under Grant No. IIS-2427770, the Department of Energy under Grant No. DE-SC0024580, and CITRIS and the Banatao Institute at the University of California.

\bibliographystyle{IEEEtran}
\bibliography{main}
\vspace{-30pt}
\begin{IEEEbiography}[{\includegraphics[width=1in,height=1.2in,clip,keepaspectratio]{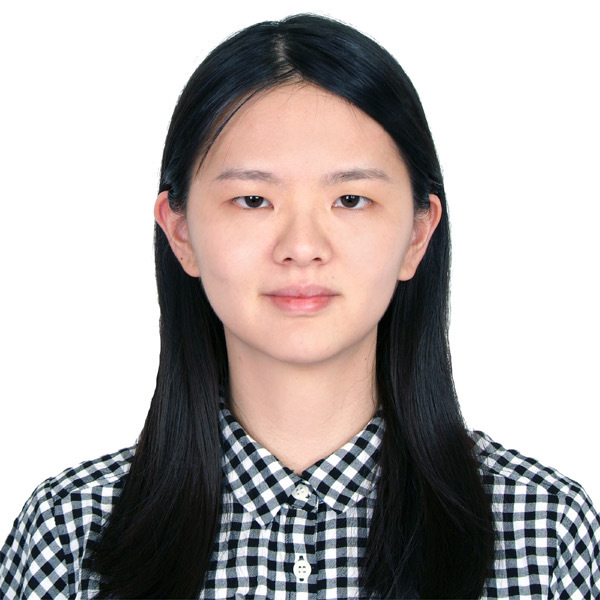}}]{Yu-Chia Huang} is a Ph.D. student at VIDI Lab, Department of Computer Science and Engineering, University of California, Davis. She obtained her bachelor's degree from National Tsing Hua University, Taiwan, and her master's degree from National Taiwan University, respectively. Her research interests include visual analytics and machine learning.
\end{IEEEbiography}
\vspace{-35pt}
\begin{IEEEbiography}[{\includegraphics[width=1in,height=1.2in,clip,keepaspectratio]{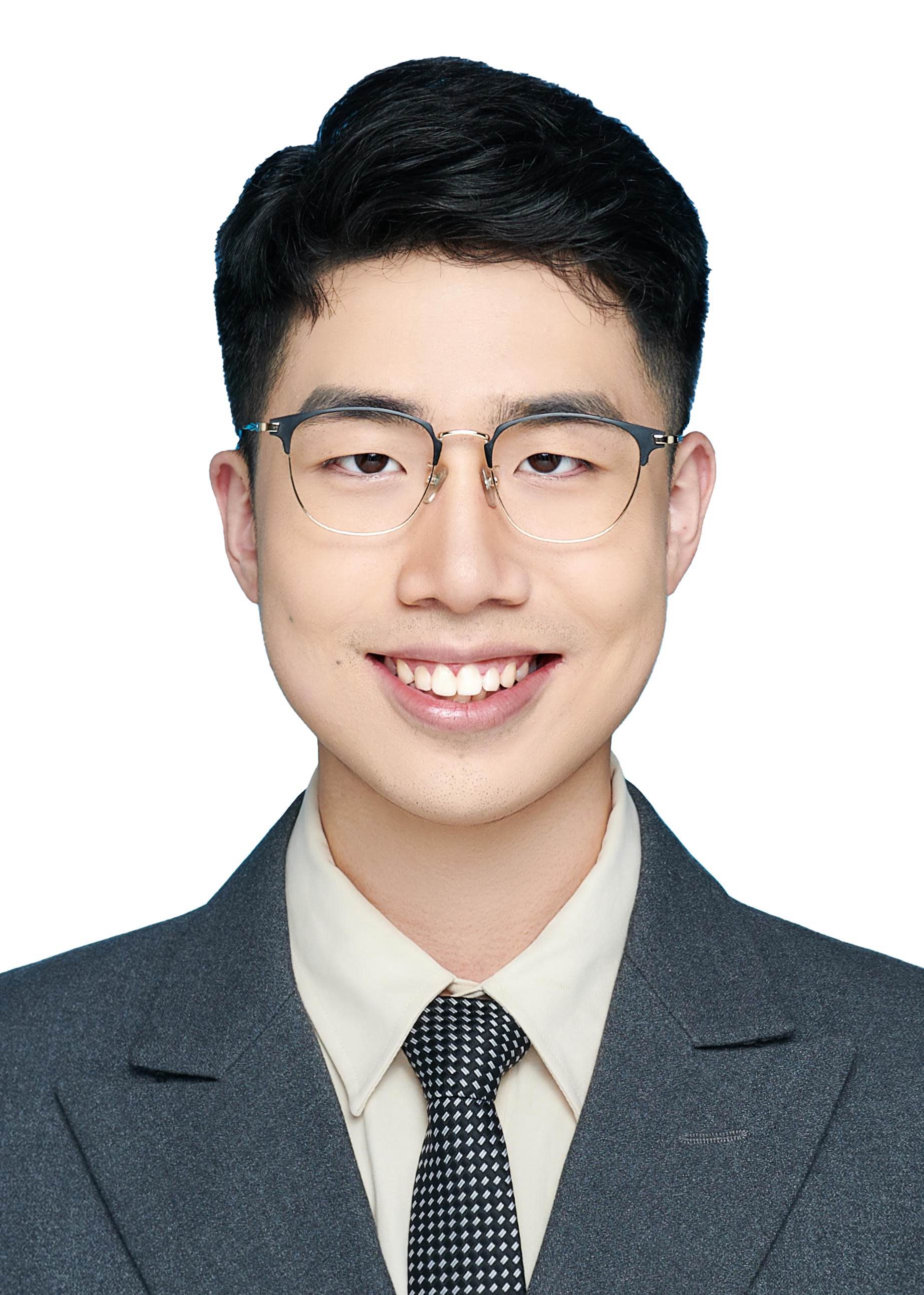}}]{Juntong Chen} received his B.Eng degree from East China Normal University, Shanghai, China, in 2022, where he is currently pursuing his Master’s degree. His research interests include large-scale data visualization, spatiotemporal data analytics, and human-computer interaction, with a focus on urban and climate data. His website is \url{https://billc.io}.
\end{IEEEbiography}
\vspace{-35pt}
\begin{IEEEbiography}[{\includegraphics[width=1in,height=1.2in,clip,keepaspectratio]{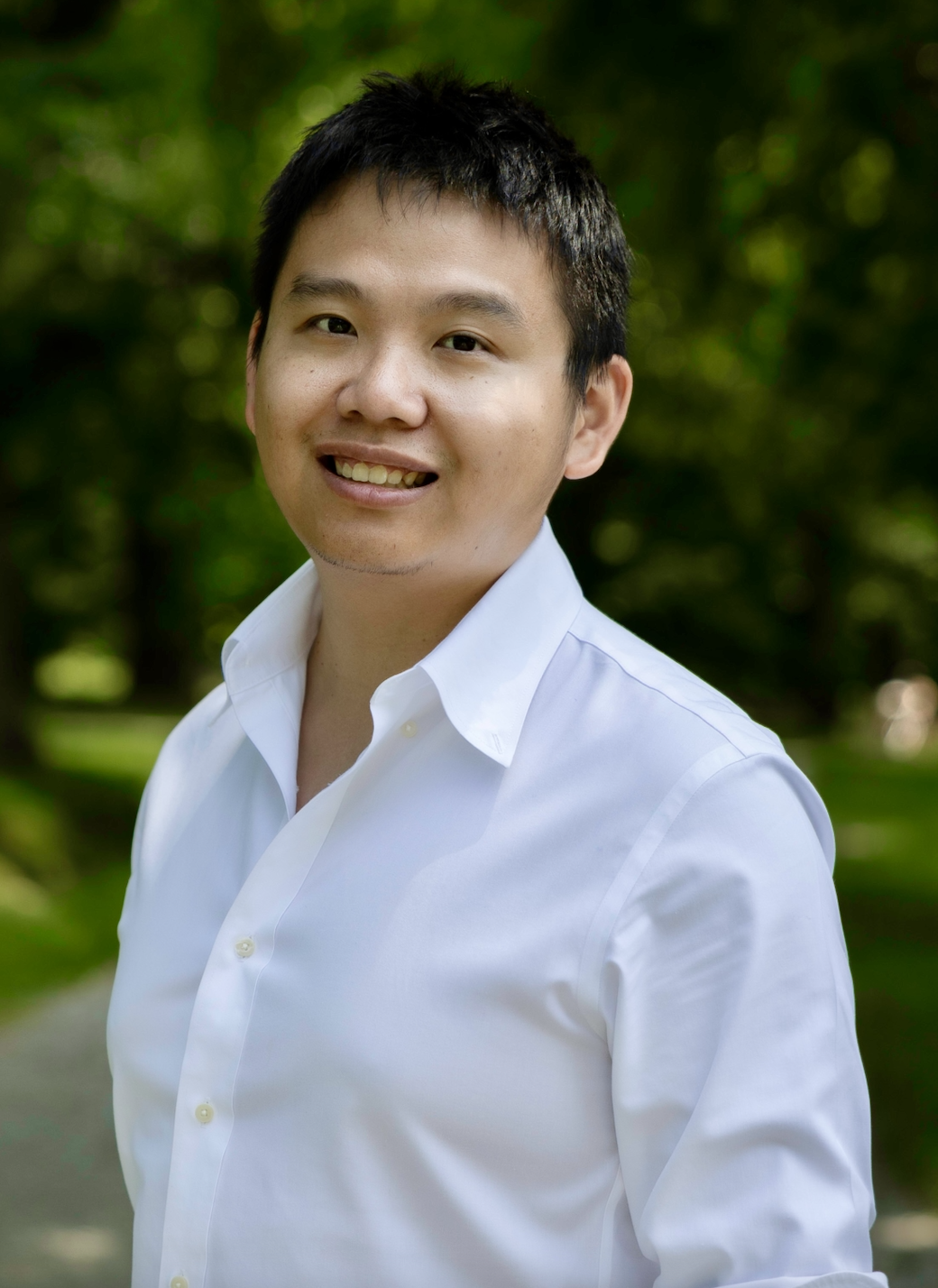}}]{Dongyu Liu} 
received the PhD degree from the
Hong Kong University of Science and Technology. He is an assistant professor with the Department of Computer Science, University of California, Davis, where he directs the Visualization
and Intelligence Augmentation (VIA) Lab. His research develops 
visualization-empowered human-AI
teaming systems for decision-making, focusing on
sustainability, and healthcare. He was a postdoctoral
associate with MIT. Website: \url{http://dongyu.tech/}.
\end{IEEEbiography}
\vspace{-35pt}
\begin{IEEEbiography}[{\includegraphics[width=1in,height=1.2in,clip,keepaspectratio]{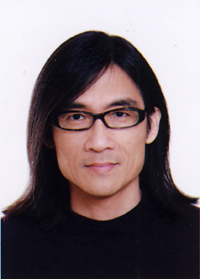}}]{Kwan-Liu Ma} (Fellow, IEEE) is a distinguished professor of computer science with the University of California, Davis. His research is in the intersection of data visualization, computer graphics, human-computer interaction, and intelligent systems. For his significant research accomplishments, he received several recognitions including being elected as ACM fellow in 2024, recipient of the IEEE VGTC Visualization Technical Achievement Award in 2013, and inducted to IEEE Visualization Academy in 2019. 
\end{IEEEbiography}

\clearpage
\appendices
\section{Scalability experiment}

To assess the scalability of our method, we compared it against a traditional shapelet extraction algorithm implemented in the \texttt{pyts} library. On a synthetic univariate time series dataset with $N=1000$ samples and $m=100$ time steps, the \texttt{pyts} algorithm required over two hours ($8557$ seconds) to extract shapelets. Moreover, the runtime of this algorithm increases exponentially with respect to both the number of samples and the length of the time series, making it impractical for larger datasets.

In contrast, we designed an experiment to evaluate the runtime of our approach on synthetic datasets with varying numbers of samples ($N \in \{1000, 2000, 5000\}$) and time series lengths ($m \in \{100, 300, 500, 1000\}$). For each configuration, we extracted 50 shapelets and set the sliding window step size to $0.01 \times m$. 

Fig.~\ref{fig:time_comp} presents the runtime of our method across varying numbers of samples and time steps. As expected, the runtime increases with both the number of samples and the sequence length. However, the increase is gradual and exhibits non-exponential growth across all settings. Notably, even for the largest dataset ($N=5000$, $m=1000$), the total runtime remains under 1000 seconds, demonstrating the scalability of our approach. In contrast to traditional shapelet-based methods, which exhibit exponential time complexity, our method remains computationally feasible as data size grows. 

As a reference for practical scenarios, Table~\ref{tab:real-world} reports the execution time for the real-world datasets used in our quantitative evaluation. Most of these datasets have relatively small input sizes, and each converges at a different rate depending on its characteristics.

\begin{figure}[htbp]  
    \centering
    \begin{tikzpicture}
        \begin{axis}[
            legend pos=north west,
            width=0.9\linewidth,
            height=0.9\linewidth,
            ylabel=Time (sec),
            xlabel=Input Time Steps,
            xmin=0, xmax=1000, 
            ymin=0, ymax=1000,
            xtick={0, 250, ..., 1000},
            ytick={0, 250, ..., 1000}
        ]
            \addplot[smooth,mark=*,Maroon] plot coordinates {
                (100,112.7)
                (300, 104.86)
                (500, 168.29)
                (1000, 191.74)
            };
            \addlegendentry{N=1000}
            \addplot[smooth,mark=*, NavyBlue] plot coordinates {
                (100, 212.11)
                (300, 208.55)
                (500, 300.2)
                (1000, 370.6)
            };
            \addlegendentry{N=2000}
            \addplot[smooth,mark=*, OliveGreen] plot coordinates {
                (100, 435.93)
                (300, 502.41)
                (500, 820.68)
                (1000, 923.23)
            };
            \addlegendentry{N=5000}
        \end{axis}
    \end{tikzpicture}
    \label{fig:time_comp}
    \caption{Runtime of our joint-learning framework on synthetic univariate time series datasets with varying sequence lengths ($m = 100$, $300$, $500$, $1000$) and sample sizes ($N = 1000$, $2000$, $5000$).}
\end{figure}
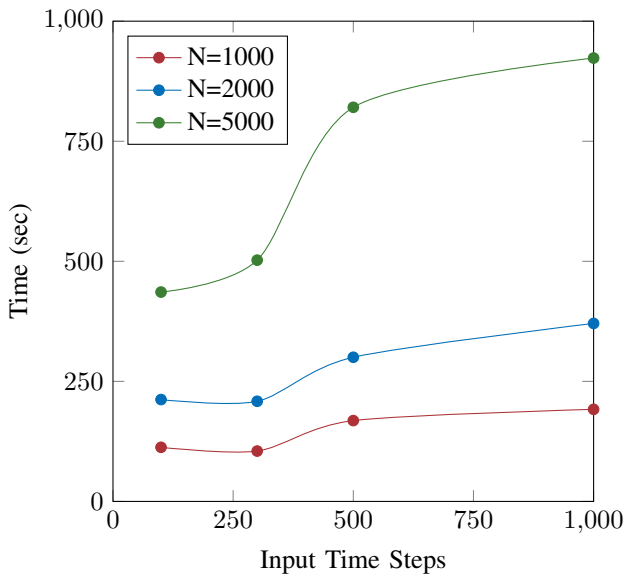
\begin{table}[!htp]
    \centering
    \begin{tabular}{c c c}
        \hline
        ECG5000-binarey & ECG200        & SonyAIBORobotSurface1 \\
        24.02           & 13.78         & 16.91                \\
        \hline
        BirdChicken     & Strawberry    & SonyAIBORobotSurface2 \\
        15.41           & 32.02         & 17.59                 \\
        \hline
        Preterm         & GunPoint      & Coffee\\
        25.10           & 25.94         & 17.34\\
        \hline
    \end{tabular}
    \caption{Training times (in seconds) for real-world datasets. The number of epochs required for convergence is different for each dataset.}
    \label{tab:real-world}
\end{table}

\section{Ablation Study}
We conducted an ablation study to evaluate the impact of key hyperparameters on the performance of the joint training framework.

\subsection{Experimental Setup}
Experiments were conducted on the \textit{Strawberry} dataset, a food spectrograph dataset used to classify strawberry and non-strawberry samples. The data were collected using Fourier Transform Infrared (FTIR) spectroscopy with Attenuated Total Reflectance (ATR) sampling. The dataset contains 983 samples, each represented as a univariate time series of length 235.

All experiments were performed under the same computational environment. Each reported result is the average over 10 independent training runs using the full training pipeline.

\subsection{Results}
Table~\ref{tab:ablation} summarizes the performance under different hyperparameter configurations. Each row represents a modified setting, with omitted values indicating that the default joint training (JT) setting was used.

\vspace{3pt}
\noindent\textbf{Group A: Learning Rate.}
A smaller learning rate is essential for stable optimization. It allows the model to converge gradually and reduces the risk of getting stuck in poor local minima. Large learning rates, in contrast, often lead to unstable or suboptimal convergence.

\vspace{3pt}
\noindent\textbf{Group B: Batch Size.}
Batch size influences both the convergence dynamics and the model’s ability to generalize. We observe that the accuracy is sensitive to this parameter, with moderate batch sizes yielding more consistent performance.

\vspace{3pt}
\noindent\textbf{Group C: Number of Signatures.}
The number of signatures directly affects the model's capacity to capture class-discriminative local patterns. Too few signatures can lead to underfitting, as important temporal structures may be missed. On the other hand, using too many signatures increases the risk of overfitting and leads to higher computational cost. A balanced choice is therefore necessary to maintain both model expressiveness and generalization.

In our visualization interface, we display only the top 10 most significant signatures by default to avoid overwhelming users and to support interpretability. On the \textit{Strawberry} dataset, we found that using 30 signatures provided a good balance between performance and interpretability. This setting is consistent with our general observation across other datasets, where the optimal number of signatures typically falls between 10 and 50. In most cases, setting the number above 50 results in diminished performance, likely due to redundancy and overfitting.

\vspace{3pt}
\noindent\textbf{Group D: PIPs Rate.}
The PIPs (Perceptually Important Points) rate controls the number of candidate segments extracted during the PPSN-based signature initialization phase. A higher PIPs rate yields more candidate segments, thereby increasing the diversity of patterns considered. At the extreme, a rate of 1 means that all possible sub-segments are evaluated, which is computationally infeasible due to the exponential growth in the number of candidates.

Conversely, a very low PIPs rate may fail to capture key structural points in the time series, leading to poor signature initialization and degraded classification performance. Based on empirical analysis across all evaluated datasets, we found that a PIPs rate of 0.2 consistently provides a good balance between candidate coverage and computational efficiency. This setting generalizes well across different datasets. 
It is also important to note that the PIPs rate is not a linear proportion of the total number of time steps (i.e., a rate of 0.2 does not imply selecting 0.2$N$ points). Instead, it acts as a relative sampling parameter within the perceptual simplification algorithm, influencing the granularity of candidate segment selection.

\begin{table*}[t!]
    \centering
    \begin{tabular}{ c | c c c c c | c}
        \hline
            &   lr      & bs    & 
            {\begin{tabular}[c]{c}number of signatures\end{tabular}}  &
            pips rate & 
            {\begin{tabular}[c]{c} window step \end{tabular}} & 
            {\begin{tabular}[c]{c}testing accuracy\end{tabular}} \\
        \hline
        JT  &   0.0005  & 128   & 30                & 0.2       &  10         & 0.950 \\
        \hline
        A   &   0.001   &       &                   &           &             & 0.938 \\ 
        \hline
        \multirow{2}{*}{B}
            &           &  64   &                   &           &             & 0.917 \\
            &           &  256  &                   &           &             & 0.918 \\
        \hline
        \multirow{2}{*}{C}  
            &           &       &  10               &           &             & 0.916 \\
            &           &       &  50               &           &             & 0.920 \\
        \hline
        D   &           &       &                   &  0.1      &             & 0.922 \\
        \hline
        E   &           &       &                   &           &   5         & 0.923 \\
        \hline
        
    \end{tabular}
    \caption{Results of the ablation study. The table reports the performance of one control group (JT) and five experimental groups (A to E), each varying a specific hyperparameter: learning rate (lr), batch size (bs), number of shapelet signatures, PIPs rate for candidate extraction, and sliding window step size. Empty cells indicate that the corresponding value was kept the same as in the JT setting. Testing accuracy is reported as the average over 10 independent runs.}
    \label{tab:ablation}
\end{table*}
\begin{figure*}[hpt!]
    \centering
    \includegraphics[width=\linewidth]{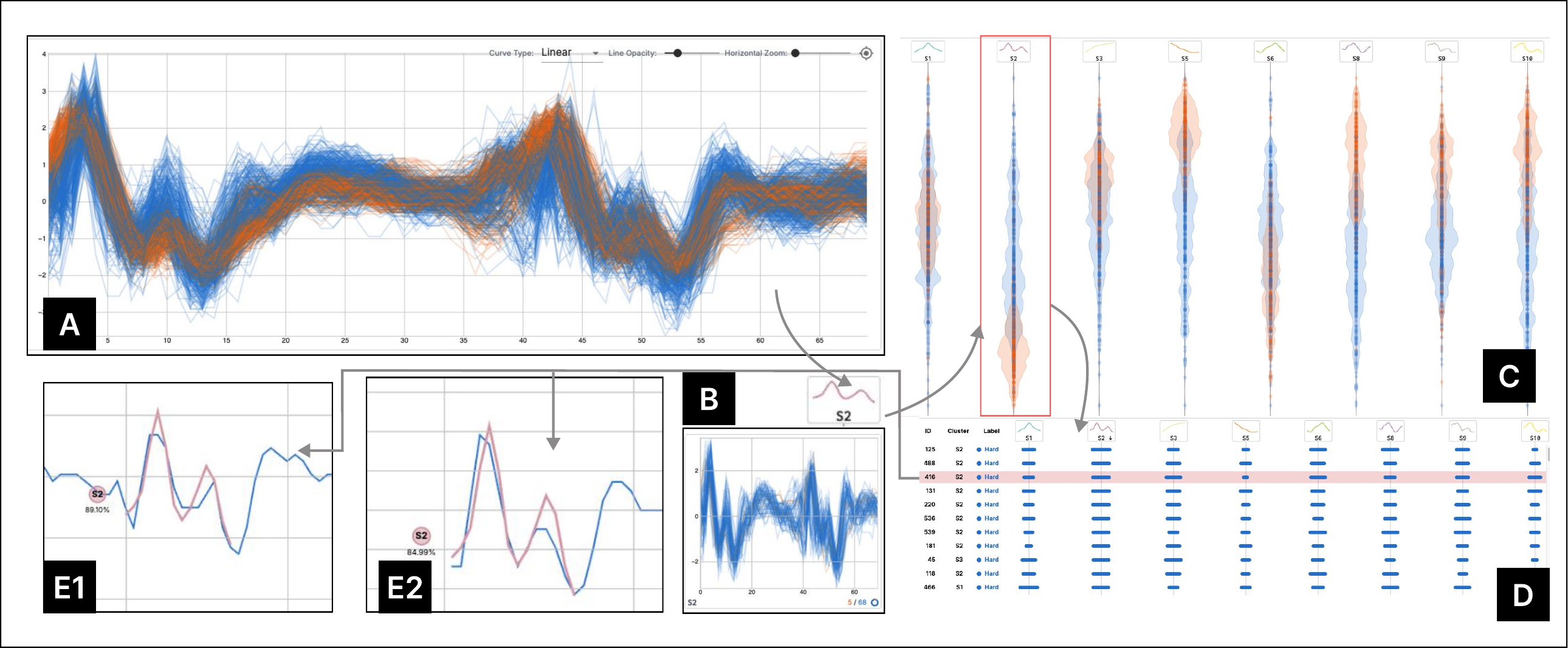}
    \caption{Usage scenario for SonyAIBORobotSurface1, a motion sensoring data for AIBO robot on cement (hard surface) and carpet (smooth surface). In the overview (A), the orange lines indicate smooth samples and the blue lines are hard samples. There are sharper peaks showing in hard samples, while not in smooth samples. In the signature list, peak-like signatures are captured (B). The relationship view shows that the distribution of signature $S2$ (C) presents separation ability, indicating a two-peak-like shape. Filtering by cluster of $S2$ (D) and ordered by score to the same signature, the signature encoding of the first few rows of the samples is similar. Selecting two samples at the top of the matrix. They are matching with signature $S2$ at a peak, suggesting that the signatrure captures the peak-like shape. That captured peak occurs at the close time steps for two aligned time series samples(E1 and E2). }
    \label{fig:robot}
\end{figure*}

\vspace{3pt}
\noindent\textbf{Group E: Sliding Window Step Size.}
The sliding window step size determines how frequently the window shifts along the time series and thus directly affects both computational efficiency and the granularity of extracted temporal information. Smaller step sizes provide higher temporal resolution but also introduce redundancy and significantly increase training time. Our results indicate that extremely fine-grained steps can lead to overfitting to local noise or fluctuations, while missing broader temporal patterns, which ultimately reduces classification accuracy.
This parameter is sensitive to the characteristics of the dataset. In our experience, time series with longer lengths ($m$) often benefit from slightly larger step sizes to avoid unnecessary redundancy. Conversely, for datasets with high-frequency temporal structure—where class-discriminative patterns tend to reoccur frequently—smaller step sizes help preserve temporal correlations and tend to yield better performance. Selecting an appropriate step size therefore requires balancing resolution and efficiency while considering both the sequence length and the underlying temporal dynamics of the data.

\section{Usage scenario: SonyAIBORobotSurface1}

\begin{figure}[hpt!]
    \centering
    \includegraphics[width=\linewidth]{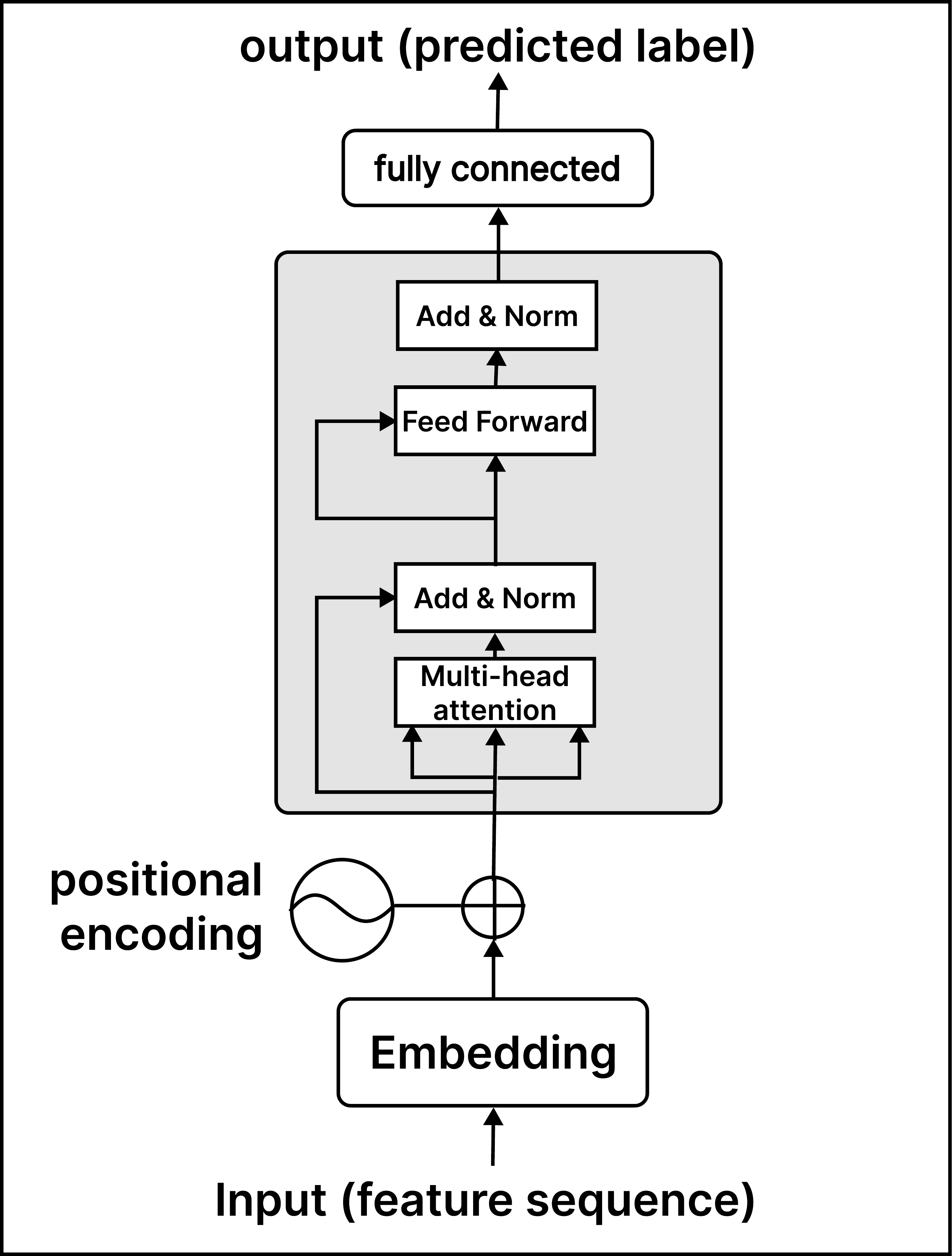}
    \caption{The architecture of the transformer in the training model.}
    \label{fig:transformer}
\end{figure}

Another usage scenario is shown in Fig.~\ref{fig:robot}, focusing on the \textit{SonyAIBORobotSurface1} dataset. This dataset contains motion sensor signals collected from an AIBO robot moving across two different surfaces: cement (hard) and carpet (smooth). As shown in the figure, our system is able to identify and interpret class-discriminative local patterns in the sensor data.
\revision{
\section{Transformer Architecture}
\sidecomment{C4}
The design of our transformer includes a regular encoder and a regular decoder with a positional encoder. The architecture of the transformer is depicted in Fig.~\ref{fig:transformer}.}

\end{document}